%% file: acl_latex.tex
\pdfoutput=1

\documentclass[11pt]{article}

\usepackage[]{acl}

\usepackage{times}
\usepackage{latexsym}
\usepackage{subfig}
\usepackage{siunitx}

\usepackage[T1]{fontenc}

\usepackage[utf8]{inputenc}

\usepackage{microtype}

\usepackage{inconsolata}
\usepackage{xspace}
\usepackage{graphicx}
\usepackage{booktabs}
\usepackage{color}
\usepackage{xcolor}  
\usepackage{pifont}  
\usepackage{caption}
\usepackage{multirow}
\usepackage{tcolorbox}
\usepackage{subfig}
\usepackage{colortbl}
\usepackage{algorithm}
\usepackage{amsmath}
\usepackage{makecell}
\usepackage{listings}
\usepackage{algpseudocode}
\newcommand{\OURS}{\textsc{EmbodiedEval}\xspace}
\newcommand{\yes}{\textcolor{green}{\ding{51}}}
\newcommand{\no}{\textcolor{red}{\ding{55}}}
\definecolor{AttrQAColor}{RGB}{230, 245, 255} 
\definecolor{SpatialQAColor}{RGB}{255, 245, 230} 
\definecolor{NavigationColor}{RGB}{240, 255, 230} 
\definecolor{ObjectInteractionColor}{RGB}{255, 230, 240} 
\definecolor{SocialInteractionColor}{RGB}{245, 240, 255} 
\title{\OURS: Evaluate Multimodal LLMs as Embodied Agents}

\author{Zhili Cheng\thanks{Core contributors, $^\ddag$Project Lead, $^{\dagger}$Corresponding Author}\hspace{0.1em}$^{\ddag}$\ \  
Yuge Tu$^{*}$ \ \
Ran Li$^{*}$ \ \
Shiqi Dai$^{*}$ \ \
Jinyi Hu$^{*\hspace{0.1em}\ddag}$\ \ 
Shengding Hu \ \ \\
\textbf{
Jiahao Li \ \ 
Yang Shi \ \
Tianyu Yu \ \
Weize Chen \ \
Lei Shi \ \
Maosong Sun$^{\dagger}$}
\\[0.5em]
Tsinghua University \ \ \\
{\texttt{\{chengzl22, hu-jy21\}@mails.tsinghua.edu.cn}} \\
}

\begin{document}
\maketitle
\begin{abstract}
Multimodal Large Language Models (MLLMs) have shown significant advancements, providing a promising future for embodied agents.
Existing benchmarks for evaluating MLLMs primarily utilize static images or videos, limiting assessments to non-interactive scenarios. 
Meanwhile, existing embodied benchmarks are task-specific and not diverse enough, which do not adequately evaluate the embodied capabilities of MLLMs.
To address this, we propose \OURS, a challenging and comprehensive benchmark to evaluate MLLMs' interactive capabilities in embodied tasks within a unified simulation and evaluation framework tailored for MLLMs.
We evaluate the state-of-the-art MLLMs on \OURS and find that they have a significant shortfall compared to human level on embodied tasks. Our analysis demonstrates the limitations of existing MLLMs in embodied capabilities, providing insights for their future development.
\end{abstract}

\input{sec/1_intro}

\input{sec/2_related}

\input{sec/3_embodiedeval}
\input{sec/4_experiments}

\input{sec/5_analysis}

\input{sec/6_conclusion}

\newpage

\section{Limitations}

To ensure the quality of the evaluation set, the verification process is time-consuming and involves checking each scene, task, and correctness individually. As a result, our evaluation set contains 327 test cases. We will incorporate more cases in future research.

\section{Potential Risks}
While we aim to advance the capabilities of MLLMs as interactive embodied agents, there are inherent risks that must be acknowledged. One potential risk is the over-reliance on MLLMs for decision-making in critical scenarios, which could lead to biased outcomes due to the models' limitations in understanding contextual nuances. Additionally, the deployment of such advanced systems in real-world environments raises concerns about privacy and data security, as these models often require substantial amounts of personal and environmental data to function effectively. 

\bibliography{custom}

\appendix

\input{sec/appendix}

\end{document}

%% file: sec/1_intro.tex
\begin{figure}[t]
  \centering
  \includegraphics[width=\linewidth]{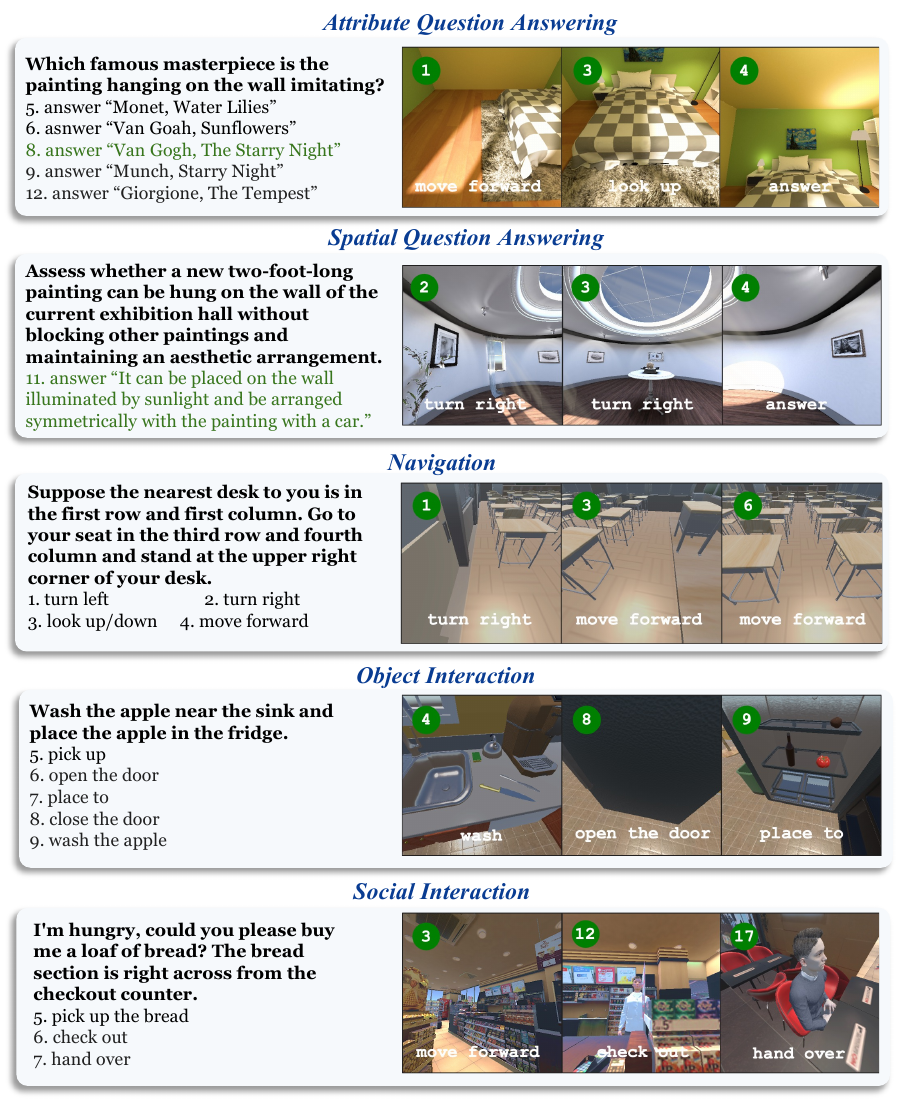}
  \caption{Examples of the five task categories and performance overview of \OURS. The embodied agent powered by MLLMs is required to finish the given task in a 3D simulation environment.}
  \vspace{-1.5em}
  \label{fig:main}
\end{figure}

\section{Introduction}
\label{sec:introduction}

In recent years, Multimodal Large Language Models (MLLMs)~\citep{openai2023gpt4v, team2023gemini, liu2024visual} have demonstrated strong capabilities in understanding and reasoning across vision and language tasks.
With the rapid development of MLLMs, a rich set of benchmarks~\citep{yue2023mmmu, MMBench,fu2023mme,li2023seed} has been developed.
Beyond these basic tasks that focus on non-interactive visual scenes, researchers are actively trying to expand MLLMs as embodied agents in interactive environments, which require the model to interpret multimodal inputs into 
actions~\citep{ahn2022can,driess2023palm,mu2024embodiedgpt}. To accomplish this, MLLMs are expected to integrate a multitude of capabilities that enable them to interact effectively with the environment, including ego-centric perception~\citep{cheng2024egothink}, visual grounding~\citep{anderson2018vision, zhang2024taskorientedsequentialgrounding3d}, spatial reasoning~\citep{chen2024spatialvlm} and episodic memory \citep{majumdar2024openeqa}.

\begin{table*}[t]
\centering
\scalebox{0.84}{
\begin{tabular}{l|cccccccc}
\toprule
Benchmark               & Scene. & Task. & Disc. & Ego.             & Nav. & Obj. & So. & Ans.   \\ \midrule
MME~\citep{fu2023mme} &-&\yes&\yes&\no    &\no&\no&\no&\yes \\
EgoPlan etc.~\citep{chen2023egoplan,egothink} &-&\yes&\yes&\yes     &\no&\no&\no&\yes           \\
OpenEQA~\citep{majumdar2024openeqa}  &-&\yes&\yes&\yes        &\no&\no&\no&\yes \\
EQA etc.~\citep{das2018embodied, yu2019multi, tan2023knowledge} &\no&\no&\yes&\yes         &\yes&\no&\no&\yes \\
ALFRED~\citep{shridhar2020alfred}   &\no&\no&\no&\yes     &\yes&\yes&\no&\no   \\
BEHAVIOR\citep{srivastava2022behavior} &\no&\yes&\no&\yes          &\yes&\yes&\no&\no \\ 
EQA-MX~\citep{islam2024eqamx}   &\no&\no&\yes&\yes    &\no&\no&\yes&\yes     \\

\midrule

\textbf{\OURS} &\yes&\yes&\yes&\yes         &\yes&\yes&\yes&\yes              \\ \bottomrule

\end{tabular}}
\caption{
Comparison of \OURS with previous benchmarks. The abbreviations in the table headers, from left to right, represent: \textbf{Scene} diversity (beyond household scenes), \textbf{Task} diversity (beyond task templates), \textbf{Disc}rete action space (for MLLMs evaluation), \textbf{Ego}centric vision, \textbf{Nav}igation involved, \textbf{Obj}ect interaction involved, \textbf{So}cial interaction involved, and \textbf{Ans}wering questions involved.
}
\label{tab:benchmark}
\vspace{-1em}
\end{table*}

However, the comprehensive evaluation of MLLMs in embodied tasks remains largely unexplored. First, existing benchmarks for embodied tasks lack diversity in both tasks and scenes. For instance, ALFRED~\citep{shridhar2020alfred} includes just seven predefined tasks (e.g., "pick and place") within four room types.
Second, several benchmarks~\cite{anderson2018vision, qi2020reverie} impose rigid input-output formats, e.g. 3D points, making it inefficient or even infeasible to evaluate mainstream MLLMs.
Third, certain benchmarks~\citep{embodiedagentinterface,liu2023agentbench,jia-etal-2024-langsuit} try to evaluate LLMs' embodied performance by representing environments with textual descriptions, relying heavily on text-based states. This downplays critical embodied skills such as visual grounding and spatial reasoning, which are essential for real-world interaction.

To address this gap, we introduce the first comprehensive benchmark for evaluating MLLMs' embodied capabilities in interactive environments. The key features of \OURS are as follows: (I) \textbf{Diverse Interactions.} \OURS provides a simulation framework that supports a wide range of interactions with objects and humans in realistic 3D environments. Agents need to interact with the environment to gather information or alter its state to complete tasks. The ego-centric visual information will serve as input to the MLLMs to make the decision. (II) \textbf{Diverse Tasks.} Unlike previous work that relied on predefined task templates, our tasks are systematically generated and carefully selected to ensure both high quality and diversity. \OURS includes novel tasks that assess a broad spectrum of abilities, enabling a more comprehensive evaluation of the model's capabilities. (III) \textbf{Diverse Scenes.} Our scenes offer significant diversity in terms of objects and spaces, encompassing house rooms, large residences, and public areas such as gyms, stores, and offices. This variety helps minimize the impact of specific scene types on the model's generalization. 

Experiment results on \OURS reveal that mainstream MLLMs largely fall short of human-level performance on embodied tasks. Model performance varies widely across different task categories, with a notable drop in spatial and long-horizon tasks. \OURS provides insights for further improvements in MLLMs's capability in grounding, spatial reasoning, planning, and exploration.

%% file: sec/2_related.tex
\section{Related Works}


\textbf{Multimodal Large Language Models.}
By connecting vision modules with LLMs, LLaVA~\citep{liu2024visual} pioneers research in MLLMs through visual instruction tuning. Many work further improves the MLLMs from various aspects, including detailed captioning~\citep{chen2023sharegpt4v}, trustworthy response~\citep{yu2023rlhf, yu2024rlaifv}, multilingual capabilities~\citep{hu2024large}, visual grounding~\citep{peng2023kosmos, you2024ferret} and video understanding~\citep{lin2023videollava, liu2024nvila}.

\textbf{Evaluation for MLLMs.}
Mainstream benchmarks for MLLMs mainly focus on perception and cognitive evaluation~\citep{fu2023mme, MMBench, yue2023mmmu,fu2024videomme} and some benchmarks focus on more challenging tasks~\citep{lu2024mathvista, he2024olympiadbench, textvqa, ocrbench, yang2024thinking, yue2023mmmu}.
Additionally, certain benchmarks~\citep{fan2019egovqa, chen2023egoplan, egothink, majumdar2024openeqa,szot2023large} have been designed to evaluate the egocentric capabilities of MLLMs using egocentric images or videos.
However, these benchmarks use static question-answering pairs without interacting with environments.

\begin{figure*}[t]
  \centering
  \includegraphics[width=0.98\linewidth]{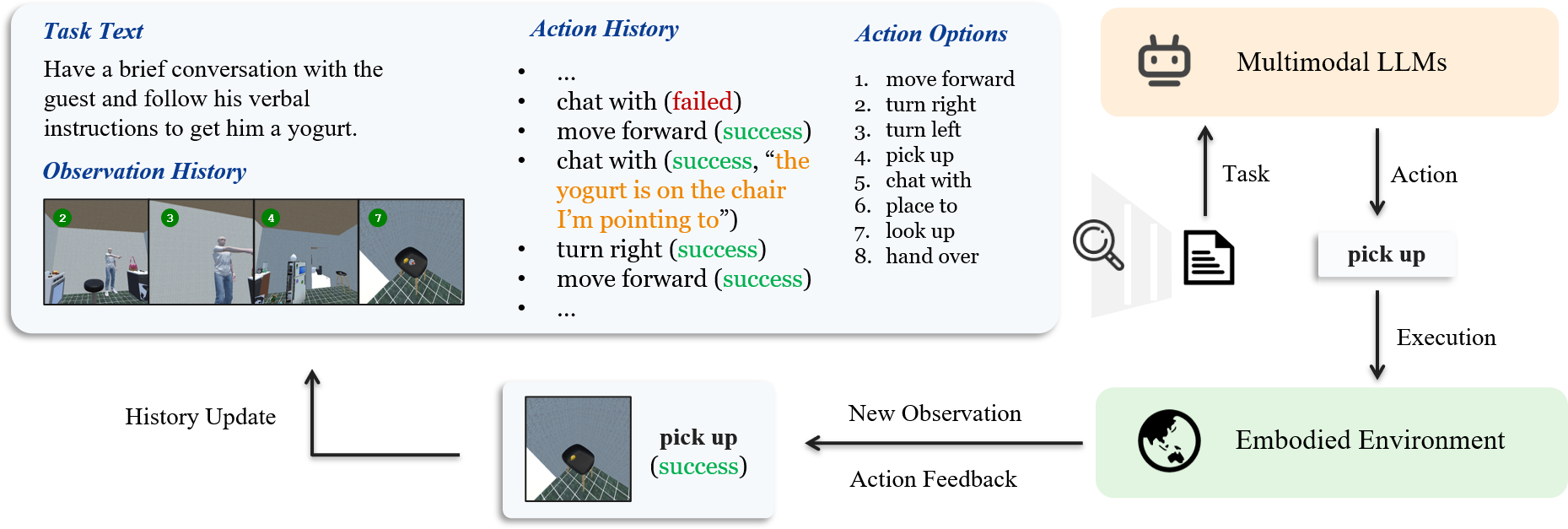}
  \caption{The evaluation process of \OURS. The task description and ego-centric observation history will be input for the model. The environment will respond to the action from the model output with a new observation.}
  \label{fig:process}
  \vspace{-1em}
\end{figure*}

\textbf{Benchmarks for Embodied Agents.}
Existing benchmarks or datasets for embodied agents cover several areas such as embodied question answering~\citep{das2018embodied,yu2019multi, tan2023knowledge, ren2024explore, islam2024eqamx, dorbala2024s, gordon2018iqa}, navigation~\citep{anderson2018vision, jain2019stay, ku2020room, zhu2021soon, qi2020reverieremoteembodiedvisual, ma2024doze, khanna2024goat} and object interaction~\citep{shridhar2020alfred, batra2020rearrangement, weihs2021visual, kant2022housekeep, misra2018mapping,srivastava2022behavior,li2023behavior}.
However, existing embodied benchmarks are limited in task variety, lacking comprehensive assessments of navigation, object interaction, and question-answering. They rely on predefined task templates, failing to adequately capture the wide spectrum of embodied capabilities. Additionally, the task-specific observation spaces and continuous action spaces in many benchmarks are inadequate for effectively evaluating MLLMs.
We summarize the comparison between \OURS and other representative benchmarks in Table~\ref{tab:benchmark}.

%% file: sec/3_embodiedeval.tex
\section{EmbodiedEval}

In this section, we introduce the evaluation process and data collection process of \OURS.

\input{sec/3_1_evaluation_framework}

\input{sec/3_3_dataset_construction}
\subsection{Dataset Statistics}
\label{Dataset Statistics}

We summarize the statistics of \OURS in Figure~\ref{fig:scene_word_statistics}. \OURS consists of 328 tasks in 5 categories across 125 unique scenes, 575 predicate instances, and 1533 varied options including 1213 textual answers and 320 interactions. Each episode requires 10.72 steps on average based on expert demonstrations.
Task descriptions average 16.09 words in length, while options average 5.72 words. The left of Figure~\ref{fig:scene_word_statistics} shows the distribution of the task across 5 task categories and 4 scene sources, the middle presents a visualization of frequent words categorized by grammatical type. See more task samples in Table~\ref{tab:task_cases}.

%% file: sec/3_1_evaluation_framework.tex
\begin{algorithm}[t]
\caption{\OURS Evaluation Process}
\label{alg:eval_alg}
\textbf{Input:} A Multimodal LLM $\pi$, a scene $x$, a task description $g$, an option list $\mathcal{C}=\{a_0, a_1, ..., a_n\}$, and a predicate list $\mathcal{P}$.\\
\textbf{Output:}  A boolean indicating whether the task was successful $success$.

\begin{algorithmic}[1] 
\State $o, s \gets E.reset(x)$  \Comment{$E$ is the simulator, $o$ is the visual observation, $s$ is the world state}

\State $\mathcal{M}_o \gets \{o\}$ \Comment{observation history}
\State $\mathcal{M}_a \gets \emptyset$ \Comment{action history}
\For{$i \gets 0 \text{\ to\ max\ steps} $}
    \State $a \gets \pi.predict(g, \mathcal{C}, \mathcal{M}_o, \mathcal{M}_a)$
    \State $o, s \gets E.step(a)$
    \State $\mathcal{M}_o.append(o)$
    \State $\mathcal{M}_a.append(a)$
    \State $done \gets P.judge(s)$
    \If{$done$}
        \State \Return $\text{true}$
    \EndIf
\EndFor
\State \Return $\text{false}$ \Comment{reach the max steps}
\end{algorithmic}
\end{algorithm}

\begin{figure*}[t]
  \centering
  \includegraphics[width=0.98\linewidth]{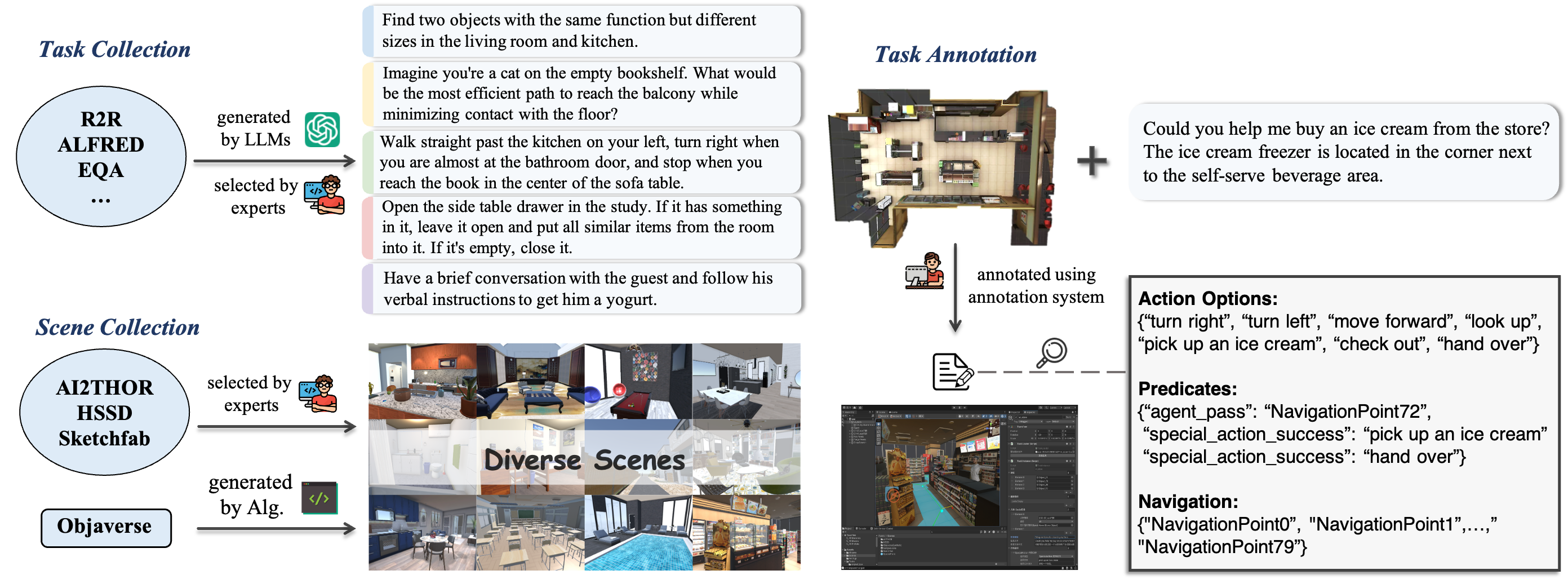}
  \caption{The dataset construction pipeline of \OURS .}
  \label{fig:pipeline}
  \vspace{-1em}
\end{figure*}

\subsection{Evaluation Formulation}
\label{sec:Evaluation Framework}
\OURS utilizes LEGENT~\citep{cheng2024legent} platform as simulator, which provide a rich and interactive 3D environment with communicable and actionable agents. We formulate the evaluate process as a decision making problem. As shown in Algorithm~\ref{alg:eval_alg} and Figure~\ref{fig:process}, an evaluation episode unfolds as follows: 
(1) The simulator initializes the 3D scene $x$. The agent $\pi$, powered by the evaluated MLLM, is positioned at a designated starting point, and the initial \textbf{first-person} visual observation $o^{(0)}$ provided by the environment is saved into the observation history $\mathcal{M}_o=\{o^{(0)}\}$. (2) At each step $i$, the agent $\pi$ chooses an action $a^{(i)}$ from a given list of options $\mathcal{C}$, which includes movement, interaction and answering, based on both the observation history $\mathcal{M}_o$ and action history $\mathcal{M}_a$. The environment executes the action, changes the state accordingly, and returns new visual observations $o^{i+1}$, along with feedback indicating whether the action was successful. The observation, action, and feedback are then appended to the observation history. (3) This process continues until either all success criteria are met, leading to task completion, or the task fails due to an incorrect answer or exceeding the maximum allowed steps. Task success is determined by the environment based on a set of predefined predicates, which maps the state of the simulation environment to a boolean value indicating success. Further details on the success criteria can be found in Appendix~\ref{sec:Success Criteria}. 

To holistically and effortlessly evaluate MLLMs' embodied capabilities with diverse tasks, rather than focusing on adapting to particular input-output requirements, we define a unified input and output space. The input space consists of textual task descriptions $g$, action option $\mathcal{C}$, and egocentric visual observation $\mathcal{M}_o$ provided by the environment, without any additional environmental state information. This design choice emphasizes visual information as it is both the most accessible and the most general medium connecting the agent to the environment. Also, visual data is the more scalable source for training multimodal foundation models comparing with low-level data. Visual observations can take the form of multiple images representing different states or a video capturing the entire process of state transitions. 

The output action space consists of movement, interaction, and answering, which varies in each task instance. For the \textbf{movement}, to make the evaluation feasible for current MLLMs, we constrain the movement space of agent on a navigation graph pre-constructed for each scene.
MLLMs are not required to make choices from a set of 3D positions, but only need to make directional decisions among navigation points. 
The details of the movement space can be found in Appendix~\ref{sec:Movement Space}

For the \textbf{interaction}, we utilize the high-level discrete interaction space. We use an open vocabulary for the actions and objects in interactions, where each action has a brief action text, operable objects, and conditions for successful interaction. For example, the ``hand over" action requires the agent to hold an object and be next to a person. 
In a given test case, several interaction actions will be involved. The details of the interaction space can be found in Appendix~\ref{sec:Interaction Space}.

For the \textbf{answering}, the agent selects an answer from a set of annotated textual responses. It can continue exploring until it believes it has enough information to make a selection. Once an answer is chosen, the task is immediately judged as correct or incorrect. The options are challenging, closely related to the context, and of high quality, as shown in Appendix~\ref{Option Cases}.



\input{table/statistics}

\subsection{Task Categories}
\label{Task Categories}

\input{table/task_samples}
\OURS defines five task categories to comprehensively assess the embodied capabilities of MLLMs:
(1) \textbf{Navigation} tasks involve coarse-grained and fine-grained natural language instructions, requiring the agent to navigate from its initial position to target locations or find specific objects.
(2) \textbf{Object interaction} tasks require agents to modify the environment through direct interaction with objects, such as moving objects, opening or closing doors and drawers, and operating electrical devices.
(3) \textbf{Social interaction} tasks involve human-agent interactions, including item delivery, perspective-taking, human feedback interpretation, and non-verbal expression comprehension.
(4) \textbf{Attribute question answering} tasks require the agent to explore the environment and answer questions related to object and scene attributes.
(5) \textbf{Spatial question answering} requires agents to answer spatial-related questions through actions and observations, such as queries about size, position, distance, layout, and spatial relationships. Each task type presents challenges that require the agent to integrate various capabilities such as grounding and reasoning. We show the samples from each category in Figure~\ref{fig:main} and more detailed examples in Appendix~\ref{Task Details}.

%% file: table/statistics.tex
\begin{figure*}[t]
    \centering
    \begin{minipage}{0.29\linewidth}
        \centering
        \includegraphics[width=0.95\linewidth]{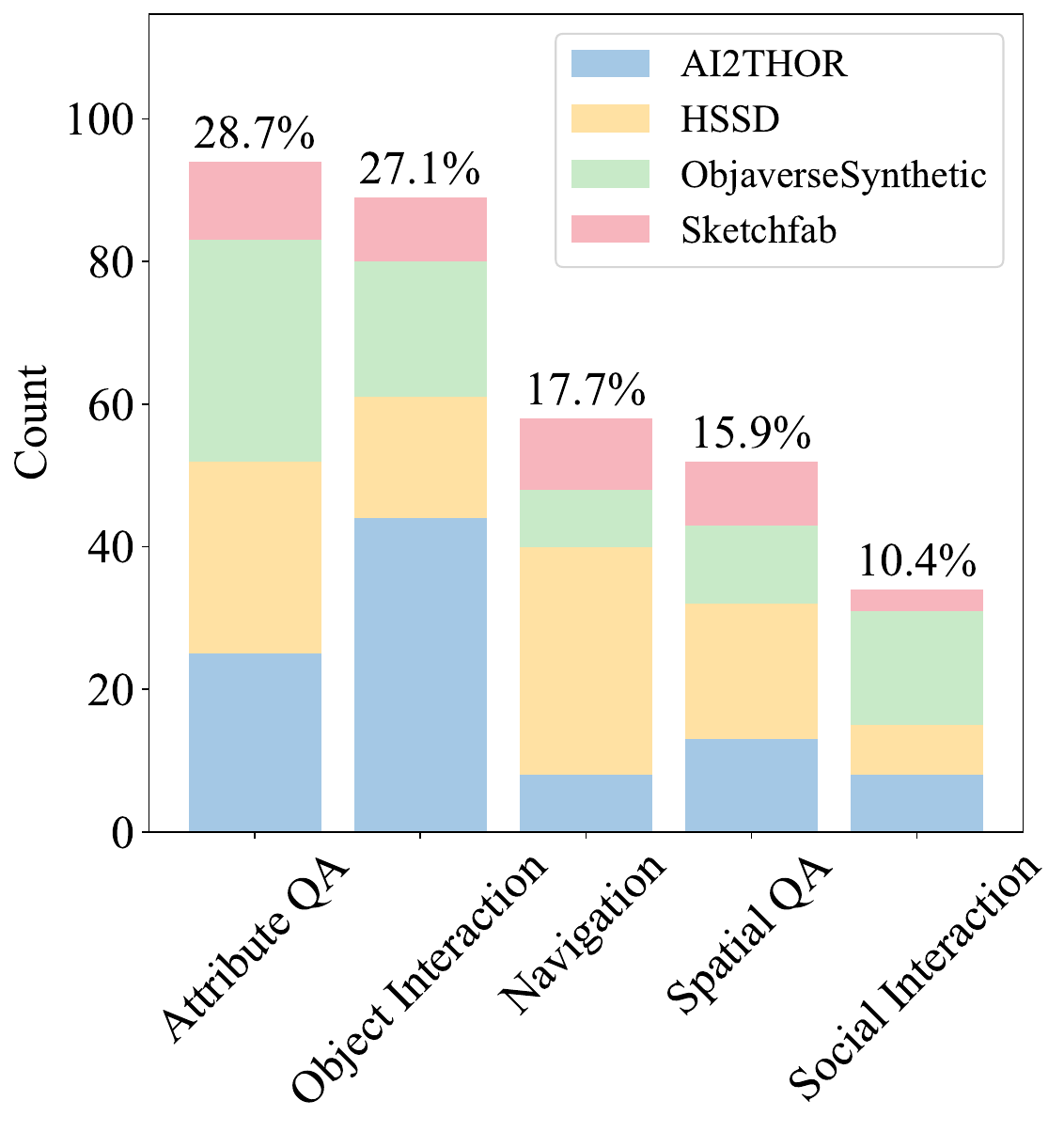}
    \end{minipage}
    \begin{minipage}{0.29\linewidth}
        \centering
        \includegraphics[width=\linewidth]{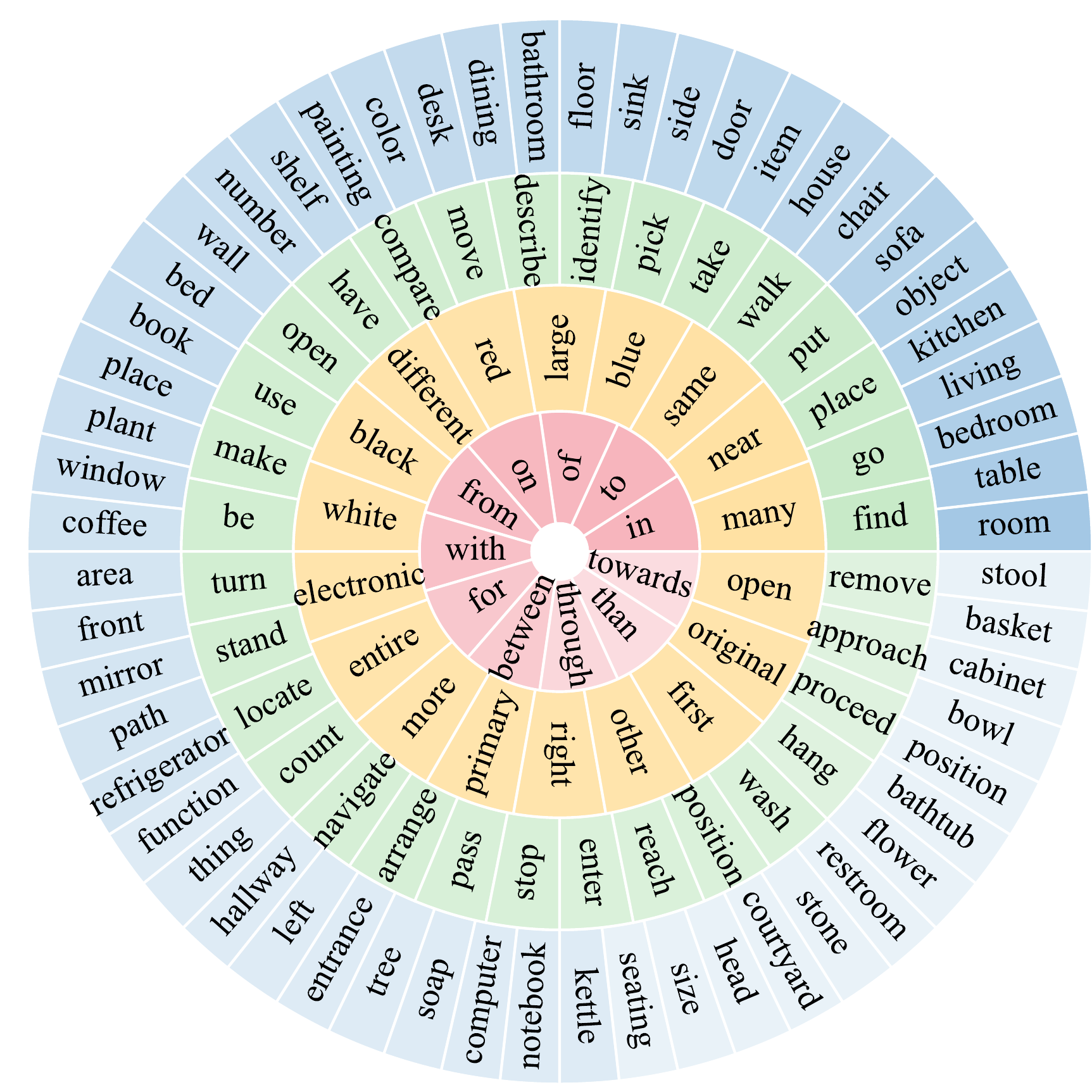}
    \end{minipage}
    \hspace{0.02\linewidth} 
    \begin{minipage}{0.28\linewidth}
        \centering
        \scalebox{0.85}{
        \begin{tabular}{ll}
            \toprule
            \textbf{Statistics} & \textbf{Number} \\
            \midrule
            Total Tasks/Predicates & 328/575 \\ 
            Total Options & 1533 \\
            \quad \textit{Textual/Interaction} & 1213/320 \\
            Total Scenes & 125 \\
            \midrule
            Average Task Length & 16.09 \\
            Average Option Length & 5.72 \\
            Average Step Required & 10.72 \\
            \bottomrule
        \end{tabular}}
    \end{minipage}
    \caption{Dataset statistics of \OURS. Left: Number of tasks by category for each scene source. Middle: Visualization of vocabulary by part of speech and word frequency.}
    \label{fig:scene_word_statistics}
    \vspace{-1em}
\end{figure*}


%% file: sec/3_3_dataset_construction.tex
\subsection{Benchmark Construction}
\label{Dataset Construction}

The construction process of \OURS consists of three parts: scene collection, task collection, and task annotation. Each sample in the dataset requires substantial effort and undergoes rigorous annotation.
Figure~\ref{fig:pipeline} illustrates the overview of dataset construction pipeline.

\input{table/main_results}

\subsubsection{Scene Collection}

We construct a diverse collection of scenes from four different sources: Objaverse~\citep{deitke2022objaverseuniverseannotated3d}, AI2THOR~\citep{ai2thor}, Habitat Synthetic Scenes Dataset (HSSD)~\citep{khanna2023habitatsyntheticscenesdataset} and Sketchfab\footnote{https://sketchfab.com}. We use Objaverse to generate indoor scenes, leveraging its extensive collection of 3D assets. First, we filter out unsuitable outdoor objects and manually review rendered assets to remove low-quality assets. See selected assets of Objaverse in Figure~\ref{fig:objaverse}. Next, we employ ChatGPT to annotate object categories with their typical room placements and functions. Using the procedural generation methods~\citep{deitke2022️}, we sequentially placed them based on their attributes. After the scene is generated, we further refine the scenes using a self-developed runtime scene editor. See more details about this synthetic process in Appendix~\ref{sec:scene_gen}. In addition, we incorporate indoor room scenes with interactive objects from AI2THOR, and some public spaces, such as stores and supermarkets, from HSSD and Sketchfab. We organize all scenes into the same format. 

\subsubsection{Task Collection}

For task collection, we first gather seed tasks for each of the five task categories from over 30 existing datasets. Using these tasks as seeds, we prompt Claude and ChatGPT to generate diverse task examples. We ask the LLMs to incorporate various capabilities, including complex grounding, episodic memory, spatial reasoning, quantitative reasoning, common sense reasoning, and planning, which resulted in many novel tasks. From this extensive task pool, we select over 300 distinct tasks as the candidate task set. Rather than asking annotators to manually write tasks for given scenes, selecting generated tasks will ensure task diversity, avoid repetition, and reduce the dependency on individual annotators' creativity or preferences.

\subsubsection{Task Annotation}
\label{sec:annotation_guideline}
After generating the task candidate set, we conduct a manual annotation to finalize each sample. First, the annotators align a suitable scene to the task from the candidate pool. Second, the annotators configure the output space, including movement, interaction, and answering, as introduced in Section~\ref{sec:Evaluation Framework}, and define the success criteria. Finally, the annotated tasks are running in the simulator to confirm that the tasks can be successfully finished. 
We recruit eight expert annotators to perform the annotations. Before beginning the annotation process, we provide systematic training on annotation requirements and system usage. To ensure high dataset quality, each annotated task is independently reviewed for correctness and accuracy by at least three evaluators. Additionally, we validate task feasibility by creating expert demonstrations for each task with expert participants and assessing human performance with non-expert participants.
See more details about the annotation process, annotation system, and quality control in Appendix~\ref{sec:appendix_annotation}.

%% file: table/main_results.tex
\begin{table*}[h!]
\centering
\captionsetup{skip=5pt}
\renewcommand{\arraystretch}{1.1}
\scalebox{0.66}{
\begin{tabular}{lccccccccccccc cc}
\toprule
\multirow{2}{*}{Model} & \multicolumn{1}{>{\columncolor{AttrQAColor}}c}{Attr. QA} & \multicolumn{1}{>{\columncolor{SpatialQAColor}}c}{Spatial QA} & \multicolumn{3}{>{\columncolor{NavigationColor}}c}{Navigation} & \multicolumn{3}{>{\columncolor{ObjectInteractionColor}}c}{Object Interaction} & \multicolumn{3}{>{\columncolor{SocialInteractionColor}}c}{Social Interaction} & \multicolumn{2}{c}{\textbf{Overall}} \\
\cmidrule(r){2-3} \cmidrule(r){4-6} \cmidrule(r){7-9} \cmidrule(r){10-12} \cmidrule(r){13-14} \cmidrule(r){15-16}
& Succ. & Succ. & Succ. & GcS & SPL & Succ. & GcS & SPL & Succ. & GcS & SPL & Succ. & GcS \\

\midrule
\texttt{Random} & 11.58 & 7.69 & 3.45 & 8.76 & 3.45 & 0.00 & 6.18 & 0.00 & 2.94 & 8.33 & 2.94 & 5.49 & 8.66 \\
\texttt{Human} & 98.95 & 92.31 & 96.55 & 97.84 & 82.28 & 97.75 & 99.44 & 90.73 & 100.00 & 100.00 & 89.96 & 97.26 & 97.94 \\
\midrule
\multicolumn{16}{c}{\textit{Proprietary MLLMs}} \\
\midrule
GPT-4o-Mini  & 31.58 & 15.38 & 27.59 & 39.51 & 15.34 & 2.25 & 17.42 & 1.50 & 5.88 & 22.06 & 2.98 & 17.68 & 25.58 \\
GPT-4o & 35.79 & \textbf{32.69} & \textbf{31.03} & \textbf{42.53} & \textbf{22.23} & \textbf{10.11} & 24.25 & \textbf{5.94} & \textbf{11.76} & \textbf{26.72} & 6.74 & \textbf{25.00} & \textbf{32.42} \\
Gemini-Flash-1.5 & 26.32 & 13.46 & 5.17 & 17.10 & 3.51 & 2.25 & 7.58 & 0.96 & 2.94 & 12.50 & 1.47 & 11.59 & 16.13 \\
Gemini-Pro-1.5 & 27.37 & 9.62 & 17.24 & 25.86 & 9.78 & 4.49 & 12.36 & 3.00 & 5.88 & 18.14 & 3.44 & 14.33 & 19.26 \\
Qwen-VL-Max & \textbf{37.89} & 17.31 & 24.14 & 30.03 & 16.87 & 7.87 & \textbf{24.91} & 5.62 & 8.82 & 22.06 & \textbf{6.86} & 21.04 & 28.07 \\
\midrule
\multicolumn{16}{c}{\textit{Open-Source Image MLLMs}} \\
\midrule
InternVL2-8B & 13.68 & 13.46 & 8.62 & 18.25 & 4.04 & 0.00 & 7.43 & 0.00 & \textbf{5.88} & 18.63 & \textbf{2.45} & 8.23 & 13.27 \\
InternVL2-40B & 14.74 & 5.77 & 6.90 & 12.93 & 3.06 & 0.00 & 7.68 & 0.00 & \textbf{5.88} & \textbf{19.12} & 2.16 & 7.01 & 11.54 \\
InternVL2-Llama3-76B & 21.05 & 13.46 & 3.45 & 9.48 & 2.18 & 0.00 & 9.08 & 0.00 & 2.94 & 13.73 & 1.14 & 9.15 & 13.79 \\
LLaVA-OneVision-7B  & 16.84 & 17.31 & 5.17 & 9.05 & 3.28 & 1.12 & 8.15 & 0.80 & 2.94 & 9.80 & 1.68 & 9.14 & 12.45 \\
LLaVA-NEXT-72B & 23.16 & 5.77 & \textbf{12.07} & 22.99 & \textbf{7.83} & \textbf{3.37} & \textbf{9.74} & \textbf{2.21} & 0.00 & 12.25 & 0.00 & 10.67 & 15.60 \\
LLaVA-OneVision-72B  & \textbf{26.32} & \textbf{19.23} & 10.34 & \textbf{23.28} & 7.53 & 1.12 & 7.81 & 1.12 & 0.00 & 12.75 & 0.00 & \textbf{12.80} & \textbf{18.23} \\
VILA-8B & 15.79 & 9.62 & 1.72 & 8.91 & 0.96 & 0.00 & 3.46 & 0.00 & 2.94 & 6.37 & 1.68 & 6.71 & 9.27 \\
VILA-40B & 17.89 & 7.69 & 0.00 & 5.75 & 0.00 & 0.00 & 3.93 & 0.00 & 0.00 & 8.58 & 0.00 & 6.40 & 9.53 \\
\midrule
\multicolumn{16}{c}{\textit{Open-Source Video MLLMs}} \\
\midrule
LLaVA-Video-7B-Qwen2  & 20.00 & \textbf{19.23} & 3.45 & 4.89 & 1.88 & 1.12 & \textbf{8.80} & 0.27 & 0.00 & 5.15 & 0.00 & 9.76 & 12.63 \\
LLaVA-NEXT-Video-32B-Qwen  & 21.05 & 7.69 & 6.90 & 14.08 & 5.34 & 0.00 & 8.61 & 0.00 & 2.94 & \textbf{12.01} & 0.98 & 8.84 & 13.39 \\
LLaVA-Video-72B-Qwen2  & \textbf{27.37} & 9.62 & \textbf{15.52} & \textbf{24.28} & \textbf{9.62} & 1.12 & 8.05 & 0.86 & 0.00 & 9.80 & 0.00 & 12.50 & \textbf{16.95} \\
Oryx-34B  & 18.95 & 3.85 & 5.17 & 13.07 & 4.89 & 1.12 & 7.02 & 1.00 & 0.00 & 8.33 & 0.00 & 7.32 & 11.33 \\
VideoLLaMA2-7B & 21.05 & 9.62 & 6.90 & 17.53 & 4.88 & 0.00 & 1.63 & 0.00 & 2.94 & 7.35 & 1.38 & 9.20 & 11.99 \\
VideoLLaMA2-72B  & \textbf{27.37} & 9.62 & 12.07 & 18.68 & 6.35 & \textbf{2.25} & 7.49 & \textbf{1.38} & \textbf{5.88} & 10.78 & \textbf{2.39} & \textbf{12.81} & 15.91 \\

\bottomrule
\end{tabular}%
}
\caption{
Results of different models on \OURS (\%). Succ., GcS, and SPL mean success rate, goal-condition success, and success weighted by path length, respectively.}
\label{tab:main_results}
\vspace{-1em}
\end{table*}

%% file: sec/4_experiments.tex
\section{Experiments}


\input{table/analysis_pic}

\subsection{Experimental Setup}
We evaluate 19 MLLMs on \OURS, including proprietary MLLMs GPT-4o/4o-Mini~\cite{openai2024gpt4o}, Gemini-Pro/Flash-1.5~\cite{team2023gemini}, Qwen-VL-Max~\cite{Qwen-VL}, open-source image MLLMs Intern-VL-8B/40B/76B~\cite{InternVL2}, LLaVA-OneVision-7B/72B~\cite{zhang2024llava}, LLaVA-NEXT-72B~\cite{zhang2024llava}, VILA-8B/40B~\cite{lin2024vila},
and open-source video MLLMs LLaVA-Video-7B/72B-Qwen2~\cite{zhang2024llava}, LLaVA-NEXT-Video-32B-Qwen, Oryx-34B~\cite{liu2024oryx}, VideoLLaMA2-7B/72B~\cite{cheng2024videollama}. 
Additionally, we introduce two special agents as reference: (1) the \textit{Random} agent, which uniformly samples actions from the option set at each step, and (2) the non-expert \textit{Human} agent, who is unfamiliar with tasks and performs actions through the simulator's user interface using the same observation and action space as models. For visual observation history, \OURS provides multiple ego-centric images at each step or, alternatively, videos capturing the entire interaction process. Proprietary and open-source image MLLMs use the former as input, while video MLLMs use the latter.
We set the maximum number of attempt steps per task as 24. The image resolution is 448×448 and the field of view is 90 degrees. All models have the temperature set to 0 during evaluation as explained in Appendix~\ref{sec:temperature}. We prompt the model to output thoughts before deciding options~\citep{yao2022react, wei2022chain}.

We evaluate agent performance using three metrics.  (1) \textbf{Success Rate (Succ.)}~\citep{liu2024agentbench,habitat_challenge,ai2thor} is the primary metric we use to measure the percentage of tasks that the agent fully completes.
(2) \textbf{Goal-condition Success (GcS)}~\citep{shridhar2020alfred, kim2024contextawareplanningenvironmentawarememory} measures partial success by calculating the proportion of goal conditions achieved, as specified by predicate functions. 
(3) \textbf{Success weighted by Path Length (SPL)} \citep{anderson2018evaluationembodiednavigationagents, batra2020objectnav} evaluates task execution efficiency in navigation and interaction tasks by considering both task success and the path efficiency relative to the expert demonstration.

%% file: table/analysis_pic.tex
\begin{figure*}[t]
    \centering
    \begin{minipage}{0.32\linewidth}
        \centering
        \includegraphics[width=\linewidth]{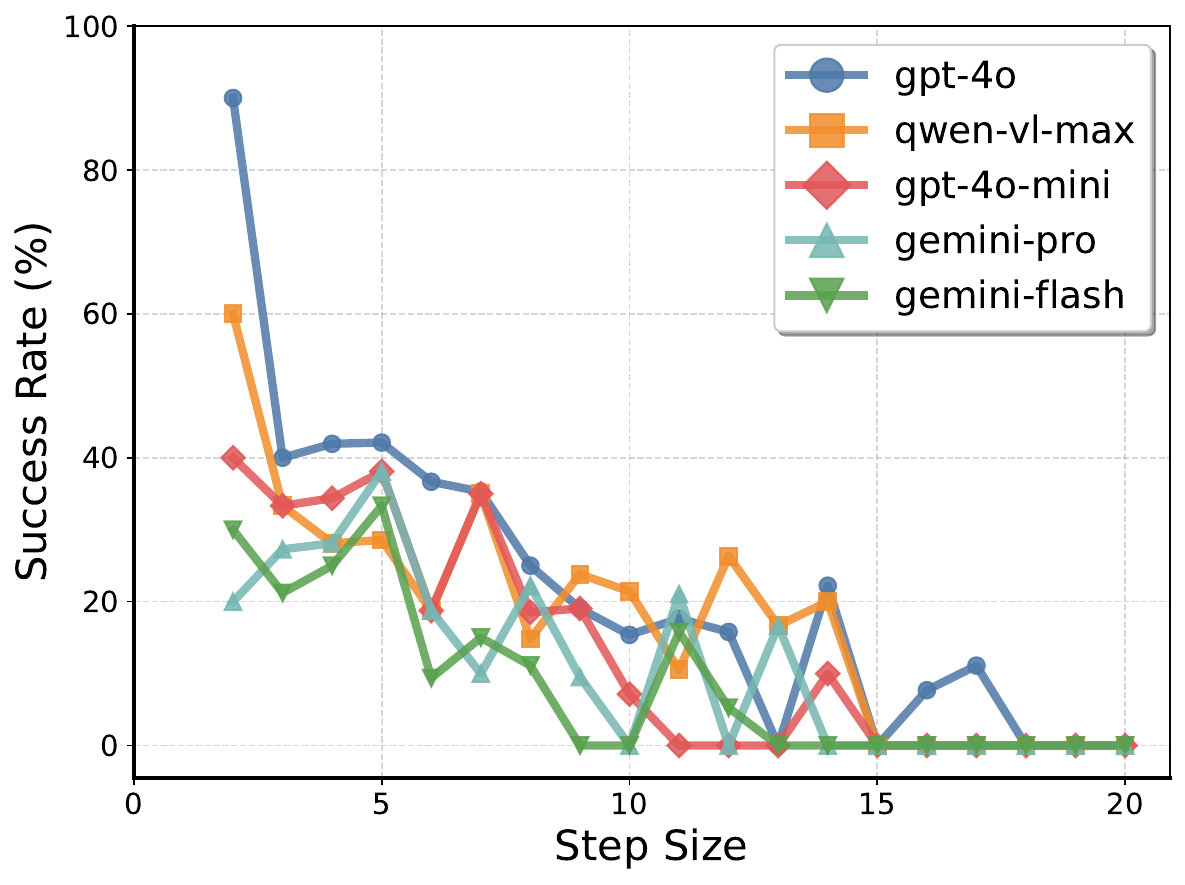}
    \end{minipage}
    \begin{minipage}{0.32\linewidth}
        \centering
        \includegraphics[width=\linewidth]{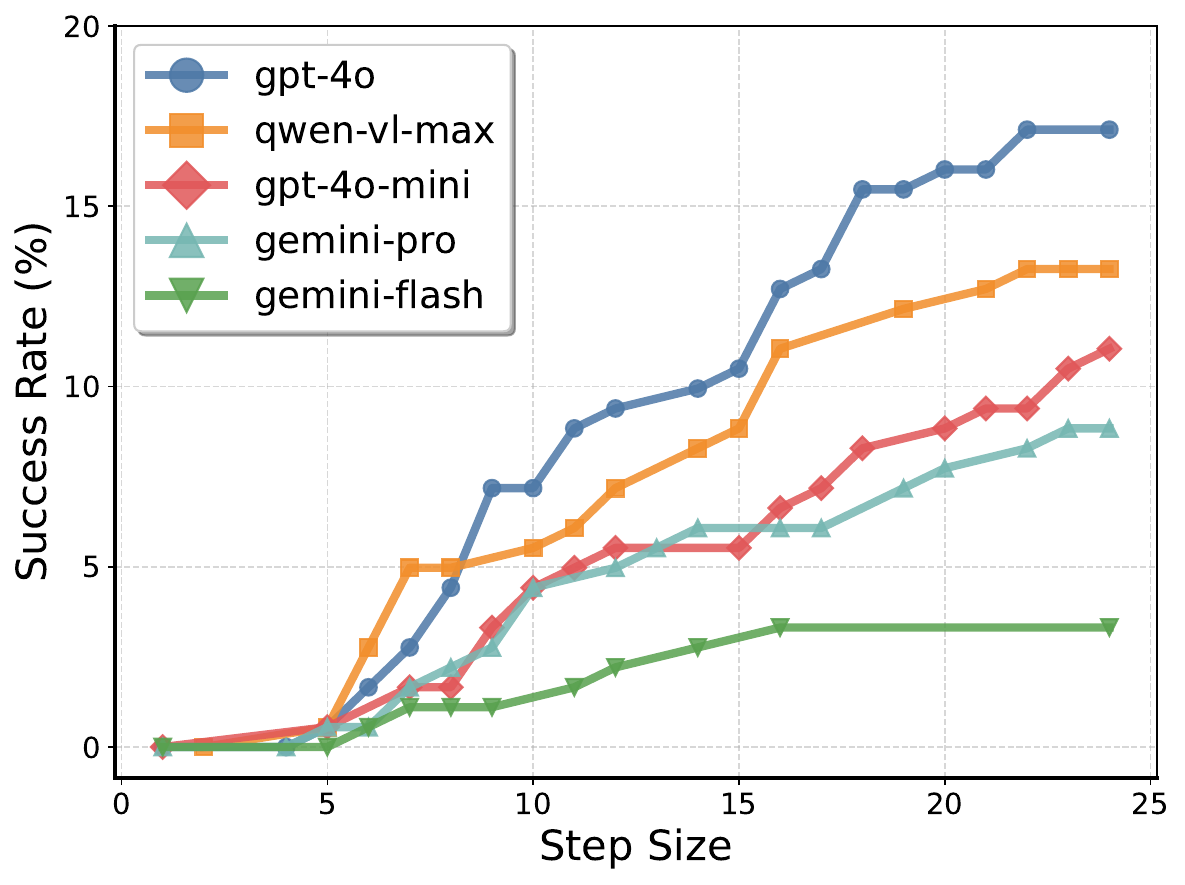}
    \end{minipage}
    \begin{minipage}{0.32\linewidth}
        \centering
        \includegraphics[width=\linewidth]{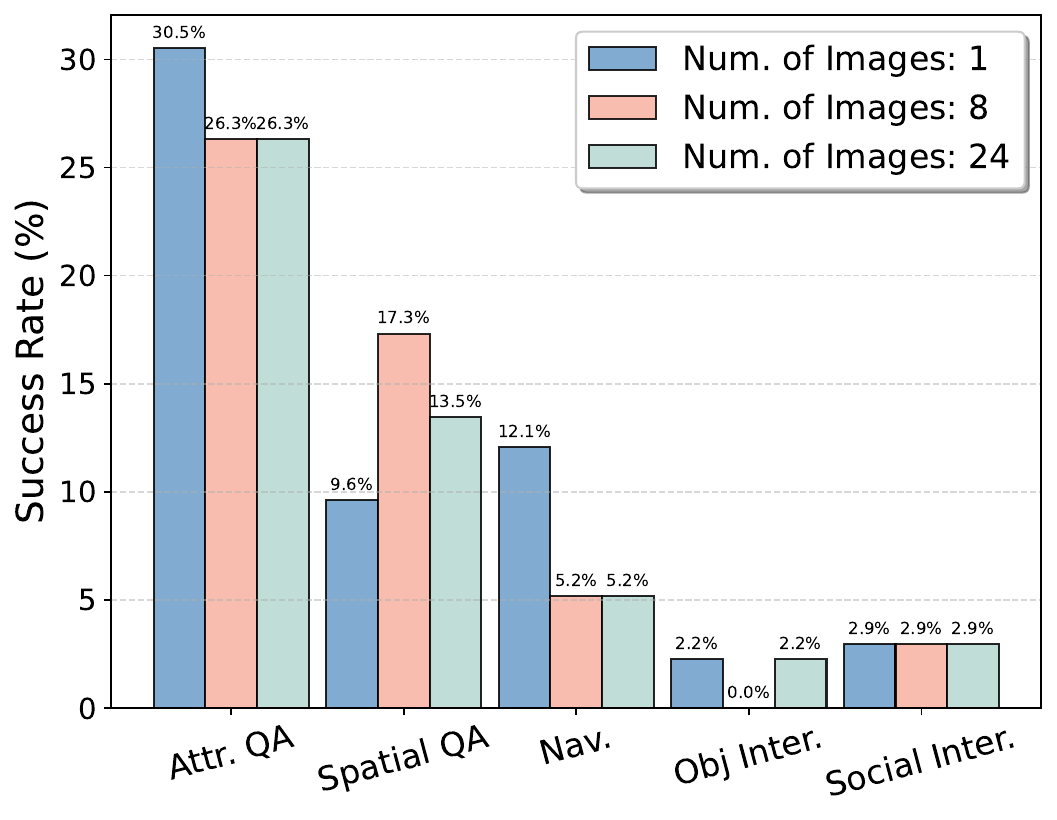}
    \end{minipage}
    \caption{Left: Success rate vs. number of steps required for the task. Middle: Success rate vs. allowed max steps. Right: The success rate of Gemini-Flash with different number of images as input across five task categories.}
    \label{fig:context_analysis}
    \vspace{-1em}
\end{figure*}

%% file: sec/5_analysis.tex
\subsection{Main Results} \label{sec:main_results}
\textbf{Performance gap between current MLLMs on Embodied Tasks with Human.}
As shown in Table~\ref{tab:main_results}, success rates across various models on \OURS remain consistently low.
The best-performing model, GPT-4o, achieves only a 25.00\% overall success rate and a 32.42\% GcS score. In contrast, non-expert humans reach a near-perfect success rate of 97.26\%, highlighting the significant challenges these models face in executing embodied tasks that humans find trivial. This performance gap is further emphasized by lower SPL scores, indicating that the models struggle to find optimal solutions. The performance open-source models show a larger performance gap. The top performing MLLM, LLaVA-OneVision-72B, achieves an overall success rate of 12.80\%, barely competitive with proprietary models. 
\textbf{Model Performance across Different Task Types.}
The results highlight a large variation in model performance across different task types. GPT-4o demonstrates relatively strong results in QA and Navigation tasks, but its performance drops notably for interaction tasks. This disparity is even more pronounced among other proprietary models. For instance, most models perform reasonably well in Attribute QA but see a sharp decline in Spatial QA that requires spatial reasoning, often halving their success rates. 
Overall, the scores for interaction tasks are consistently lower across all models, underscoring the challenge these models face in scenarios that require a deeper understanding of affordance~\citep{gibson1977theory} or social cues.
\subsection{Performance Analysis}

\textbf{Challenges in Long-Horizon Tasks.} 
We shows the trend of success rate under tasks of varying steps required for finishing. Models maintain relatively high success rate in tasks that require fewer steps but shows a decline as the task length increases. The drop in performance can be attributed to the increased complexity of longer tasks and the difficulty in handling long context. The middle of Figure~\ref{fig:context_analysis} shows the performance curve on the interaction tasks when gradually increasing the max allowed step from 1 to 25. While the success rate improves initially, the gain diminishes as the allowed steps increase, suggesting that the model struggles to manage longer histories effectively.
In the right of Figure~\ref{fig:context_analysis}, we show the performance of Gemini-Flash across different tasks with context of varying number of input images. 
Although increasing the number of images theoretically provides more historical information, the performance decreases except for spatial question answering, which benefits from the additional spatial context. This result indicates that current multimodal models still face challenges when handling multiple egocentric image inputs. 
These results highlight the difficulty of long-horizon embodied tasks, where longer sequences complicate the agent's ability to plan and act based on historical information.

\begin{table}[t]
\centering
\scalebox{0.8}{
\begin{tabular}{l S[table-format=2.2] S[table-format=2.2]}
\toprule
\textbf{Model} & {\textbf{Inter. Freq (\%)}} & {\textbf{Inter. Succ (\%)}} \\
\midrule
\texttt{Random}  & 41.06 & 2.79 \\
\texttt{Human}  & 19.44 & 96.81 \\
\midrule
GPT-4o-Mini & 26.02 & 11.34 \\
GPT-4o & 40.46 & 9.56 \\
Gemini-Pro-1.5 & 11.46 & 10.03 \\
Gemini-Flash-1.5 & 11.03 & 8.6 \\
Qwen-VL-Max & 26.33 & 8.89 \\
\bottomrule
\end{tabular}
}
\caption{Statistics of interaction tasks. \textbf{Inter. Freq} means interaction frequency and \textbf{Inter. Succ} means interaction success rate.}
\vspace{-1.5em}
\label{tab:interaction_statistics}
\end{table}

\textbf{Challenge in Interaction Tasks.}
To further analyze the low performance in interaction tasks, we show some statistics of interaction task in Figure~\ref{tab:interaction_statistics}. 
Interaction frequency measures the proportion of interaction actions among all executed actions, while interaction success rate reflects how often these actions are invoked in correct conditions, indicating the model’s affordance judgment ability. Humans can generally ensure that only necessary interactions are performed, while models exhibit varying interaction frequencies but relatively low success rates. GPT-4o achieves better performance in interaction tasks by maintaining a comparable success rate with a higher interaction frequency. Existing MLLMs need improvements in spatial perception, grounding, and affordance judgment to achieve a higher interaction success rate.

\subsection{Error Case Analysis}
\begin{figure}[t]
  \centering
  \includegraphics[width=0.95\linewidth]{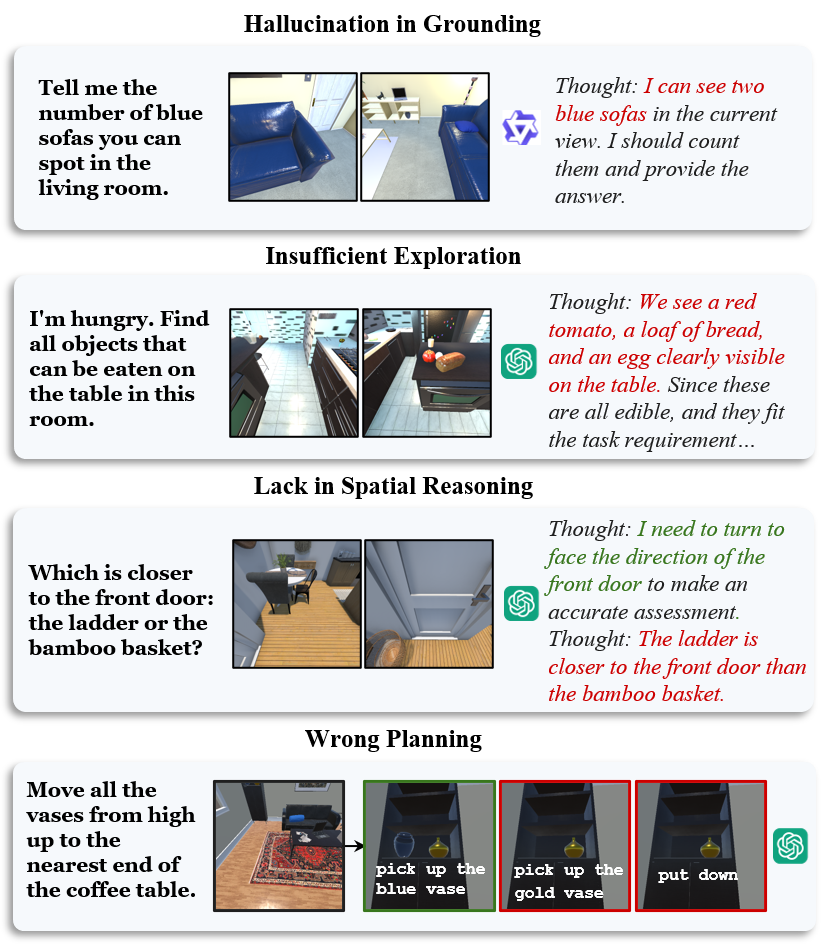}
  \vspace{-0.5em}
  \caption{Case study of common error categories. 
  }
  \label{fig:error_analysis}
  \vspace{-1em}
\end{figure}

We summarize four primary error categories in MLLM-based embodied agents: \textbf{(1) Hallucination in Grounding:} Models misperceive the environment, hallucinating nonexistent objects or overlooking present ones. For example, models may confidently describe absent items or fail to locate small objects like laptops or keys, impacting both QA (e.g., providing answers based on imagined objects) and non-QA tasks (e.g., failing to navigate to or interact with target objects). \textbf{(2) Insufficient Exploration:} 
Agents employ suboptimal exploration strategies, hindering information gathering and goal finding due to incomplete environment coverage. They are often trapped in local areas, or answer before fully exploring the environment due to overconfidence.
\textbf{(3) Lack of Spatial Reasoning:} 
Models struggle with understanding spatial relationships.
They misinterpret directional instructions (e.g.,``to my left") and face difficulties navigating between locations, even for simple tasks such as moving to or around furniture.
\textbf{(4) Wrong Planning:} Agents demonstrate poor state estimation and action planning. 
This results in random or repetitive actions, such as aimless circling or repeatedly picking up objects. They also struggle to understand the outcomes of the action and adapt after failed attempts.
Figure~\ref{fig:error_analysis} provides illustrative examples of these errors.
For more detailed examples, please refer to Appendix~\ref{sec:success_cases} and Appendix~\ref{sec:error_cases}.

\subsection{Future Improvements}
Based on the results and error analysis, there are some potential improvements for the development of MLLMs. MLLMs are primarily trained using internet data, lacking training in physical space, which is a significant difference from humans. This leads to poor spatial-related abilities, which could potentially be improved through embodied trajectory data, egocentric video data, synthetic data, and other sources. Egocentric perception and grounding in sequential images or videos should be further explored to reduce hallucination phenomena.
Since current models struggle with long-horizon tasks (even those that are just a dozen or so steps), considerable effort is needed to enable them to better understand long multimodal sequences, which is crucial for solving long-horizon visual and embodied tasks. In addition, MLLMs can also be combined with training methods like reinforcement learning to further enhance their ability to explore, reason, and recover from mistakes, building upon their foundational capabilities.

%% file: sec/6_conclusion.tex
\section{Conclusion}
In this paper, we propose \OURS, the first interactive benchmark designed for MLLMs with comprehensive embodied tasks. We provide an efficient framework to interactively evaluate the capabilities of MLLMs on embodied tasks. To ensure the accuracy, diversity, and quality of the dataset, extensive efforts are devoted to the annotation process for each task sample.

Through experiments, we find that current MLLMs perform poorly on embodied tasks. However, we believe there will be more attention to improving the embodied capabilities of MLLMs upon the general capabilities learned from universal multimodal data. We hope \OURS  can help and guide the development of MLLMs to realize their potential in embodied intelligence.


%% file: sec/appendix.tex
\clearpage

\section{Task Samples}
\label{Task Details}

In EmbodiedEval, each category of tasks has sufficient diversity to comprehensively evaluate the model. For example, different from traditional EQA task, attribute question answering tasks in \OURS encompass a more diverse range of attribute questions about objects and scenes, including but not limited to category, shape, material, color, function, state, location, existence, quantity, comparative analysis, and complex reasoning across multiple attributes and multiple objects. For interaction tasks, the tasks possibly involve multiple objects and multi-step interactions, such as using a tool to manipulate another object or rearranging items to meet specific conditions, which necessitates fine-grained movement planning, reasoning about object affordances, and understanding cause-and-effect relationships.
We selected some representative examples to illustrate the diversity of the task set in Table~\ref{tab:task_cases}.

\section{Details of Evaluation Framework}

\subsection{Movement Space}
\label{sec:Movement Space}
\begin{figure}[h]
\centering
  \centering
  \includegraphics[width=0.5\linewidth]{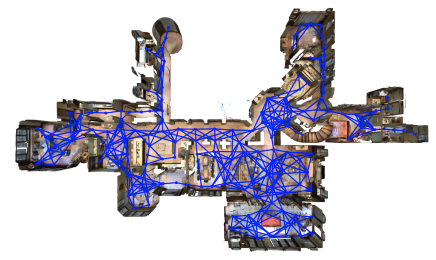}
  \hspace{-0.01\linewidth}
  \includegraphics[width=0.4\linewidth]{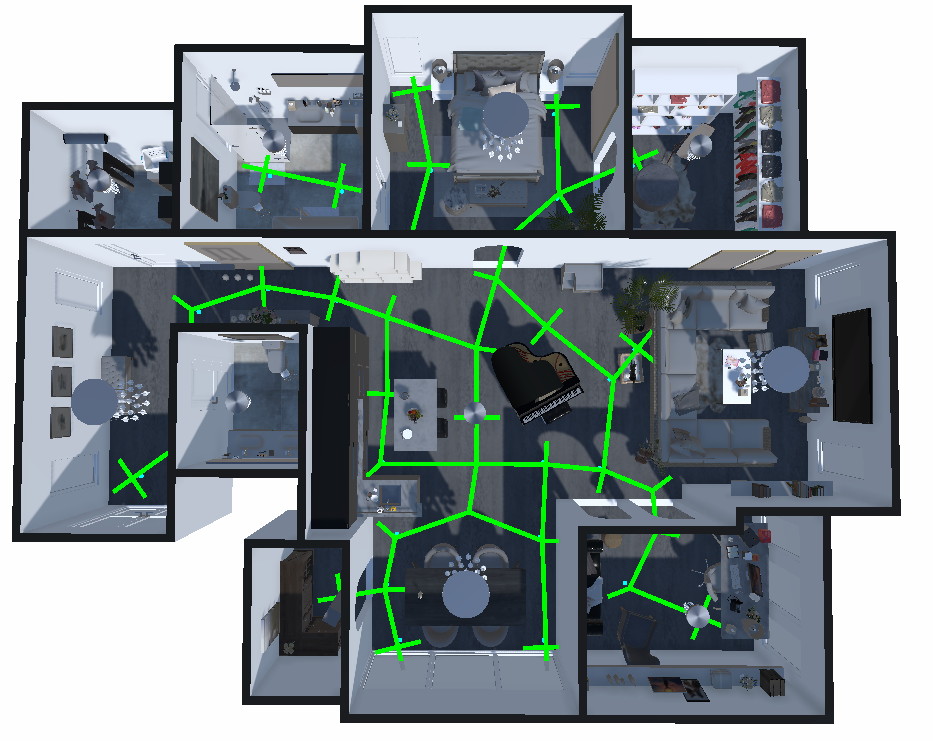}
\caption{A comparison of navigation graphs between R2R~\citep{anderson2018vision} dataset (left) and \OURS (right).}
\label{fig:navgraph}
\vspace{-1em}
\end{figure}
We use navigation graph as the movement space where the agent can rotate its view at a point or move between adjacent navigation points. Compared to continuous movement, it discretizes motion without imposing great restrictions on the high-level tasks in practice \citep{anderson2018vision}. Different from grid-world movement, this approach is more natural and adaptable to all kinds of scenes. Through sampling algorithms and manual adjustment, we constructed navigation graphs for each scene. To ensure realism, the navigation points are always walkable locations with no obstacles among them. Due to the greater diversity of our scenes and tasks compared to previous work, the density of navigation points varies based on the size of the scene and the task, ensuring that the number of steps required for tasks remains reasonable. For example, in complex interaction tasks within large scenes, the navigation points are more sparse and critical. In contrast to previous datasets, our navigation points are better organized as shown in Figure \ref{fig:navgraph}, and the connections between these points indicate clear semantics. MLLMs are not required to make choices from a set of 3D positions, but only need to make directional decisions among navigation points. Specifically, the action space consists of three types of actions: \textbf{\textit{move forward}} (moving to the facing navigation point), \textbf{\textit{turn left/right}} (rotating to face a new navigation point), and \textbf{\textit{look up/down}} (adjusting the vertical view).

\subsection{Interaction Space}
\label{sec:Interaction Space}
We follows the discrete interaction space of previous embodied AI tasks that involves object interaction such as IQA~\citep{gordon2018iqa}, CHAI~\citep{misra2018mapping}, RoomR~\citep{weihs2021visual} and OVMM\citep{yenamandra2023homerobot} rather than continous space~\citep{shridhar2020alfred,srivastava2022behavior}. This choice is based on two main considerations:
(1) In continuous spaces, interactions are tightly related to specific methods and types of embodiment,
which contradicts the goal of generality in evaluations and goes beyond the core issues of our research. (2) Due to the high complexity of continuous space, MLLMs cannot output reasonable values without being trained on specialized numerical trajectory, leading to infeasible evaluations. 
In \OURS, we use an open vocabulary for the actions and objects in interactions to make them as rich as possible. Each interaction action has a brief action text, operable objects, and conditions for successful interaction. For example, the ``pick up" action requires the target object to be within sight and very close, the ``wash" action requires the agent and the target object to be next to a sink, and the ``hand over" action requires the agent to hold an object and be next to a person. In a given test case, several interaction actions will be involved, including those necessary to complete the task and other distracting actions.
We provided more examples of the interaction space mentioned in Table~\ref{tab:action}.


\subsection{Answering Space}
\label{Option Cases}
Our answering space consists of eight annotated options by annotators. All the options are carefully written and verified to ensure that the answers are challenging, meaningful within the scene, and have a strong distractive capability. We demonstrate some examples in Figure~\ref{tab:answer options}.

\subsection{Success Criteria}
\label{sec:Success Criteria}
We automatically and accurately evaluate task completion through predicate functions. Each predicate maps the state of the simulation environment to a boolean value indicating success. For example, the \textbf{\textit{agent\_at}} predicate requires a designated navigation point as a parameter and returns true when the agent reaches this location at the end of the episode.
Beyond evaluating only the final state, \OURS also includes predicates that assess the entire process, similar to R4R~\citep{jain2019stay}. For example, the \textbf{\textit{agent\_pass}} predicate becomes true once the agent passes a specified navigation point. All the predicate are listed in Table~\ref{tab:predicates}.

A task is considered successful when all predicates evaluate to be true at the end. Consider the task ``\textit{Please go to the kitchen, then come back and tell me if there are any extra cups}". This task involves three predicates: \textbf{\textit{agent\_pass}}, \textbf{\textit{agent\_at}}, and \textbf{\textit{choose}}. These predicates verify that the agent passes through the kitchen doorway, returns to the initial position in front of the person, and selects the correct answer, respectively.

\section{Details of Task Annotation} 
\label{sec:appendix_annotation}
\subsection{Annotation Process}
A task sample includes a scene, task description, output space, and success criteria. The annotators are required to conduct the annotation as following process: (1) Select a task to annotate from the candidate tasks and choose a suitable candidate scene. Nouns, prepositions, adjectives, and other elements in the task text can be slightly adjusted to fit appropriately within the context of the scene, while keeping the core content of the task the same. Each candidate task can only be selected and used once. (2) Annotate the movement space by adjusting the navigation points in the scene. (3) For tasks involving interaction, annotate the interaction space by setting action options, including the action's text, type, and parameters. The interaction space includes the necessary action options for the task, as well as some distracting action options. For certain specific interactions, it is necessary to annotate feedback content. For example, interactions that involve asking humans require annotating the content of human responses. (4) For QA tasks, annotate the answering space by writing challenging answer options. (5) Annotate the success criteria by setting predicate functions, including the predicate's type and parameters. If the task has multiple sub-goals, the predicate function of each sub-goal should also be included. (6) Annotate the agent's initial position and orientation. (7) For social interaction tasks, annotate the initial position, orientation, and body posture of the person, including standing, sitting, lying down, and the direction of the finger pointing, and choose the person's appearance from a selection of characters from Mixamo\footnote{https://www.mixamo.com/}.
(8) Run the annotated tasks in the simulator to confirm that the tasks can be completed without any issues.

\subsection{Annotation Criteria}
\label{Annotation Criteria}
(1) All tasks must be unambiguous within the given scene.
(2) Question-answering tasks must require scene observation, with each task providing eight answer options that vary in difficulty and include misleading options to reduce the chance of guessing the correct answer. 
(3) Once a task is correctly annotated, the tasks must be executable in the simulator with a well-designed navigation graph and accurate action options. Annotators must verify task feasibility using the same observational constraints as agents.

\subsection{Annotation System}
\label{Annotation System}
To ensure both efficiency and precision in the complex annotation process, we develop an annotation system based on Unity\footnote{https://unity.com/}. The system provides comprehensive function, which encompassing scene and task import/export, flexible content viewing, visualized action space, and a guide annotation workflow that adheres to predefined guidelines:
(1) Importing and exporting scenes and tasks, allowing users to freely view the content of scenes and tasks. (2) Enforcing task annotation according to predefined guidelines. The system provides candidate lists for all types of actions and predicates and specifies he parameters that need to be annotated. (3) Generating navigation points and constructing a navigation graph with visualizations, allowing for the addition, deletion, and modification of navigation points. (4) Annotators can visually select 3D objects in the scene as parameters of interactions and predicates. Once one annotation is complete, the task file exported by the annotation system can be loaded by the simulator, starting the simulation and evaluation process.

\subsection{Quality Control}
Eight expert annotators are recruited to perform the annotations. These standard annotators are from professional data annotation companies. Before starting the annotation process, we conduct systematic training on annotation requirements and system usage.
To ensure the dataset's high quality, each annotated task is independently evaluated for correctness and quality by at least three reviewers.
There are two rounds for the annotation. In the first round, annotators primarily ensure task completeness. In the second round, expert annotators verify and refine the diversity of task objects, the accuracy and clarity of task descriptions, and task difficulty distinctions.
Furthermore, we validate task feasibility by providing expert demonstrations for each task and testing human performance with non-expert participants.

\section{Creation of Objaverse Synthetic} 
\label{sec:scene_gen}
We use a wide variety of objects from Objaverse to procedurally generate diverse scenes and further refine them through interactive scene editing.

\begin{figure*}[h]
  \centering
  \includegraphics[width=0.9\linewidth]{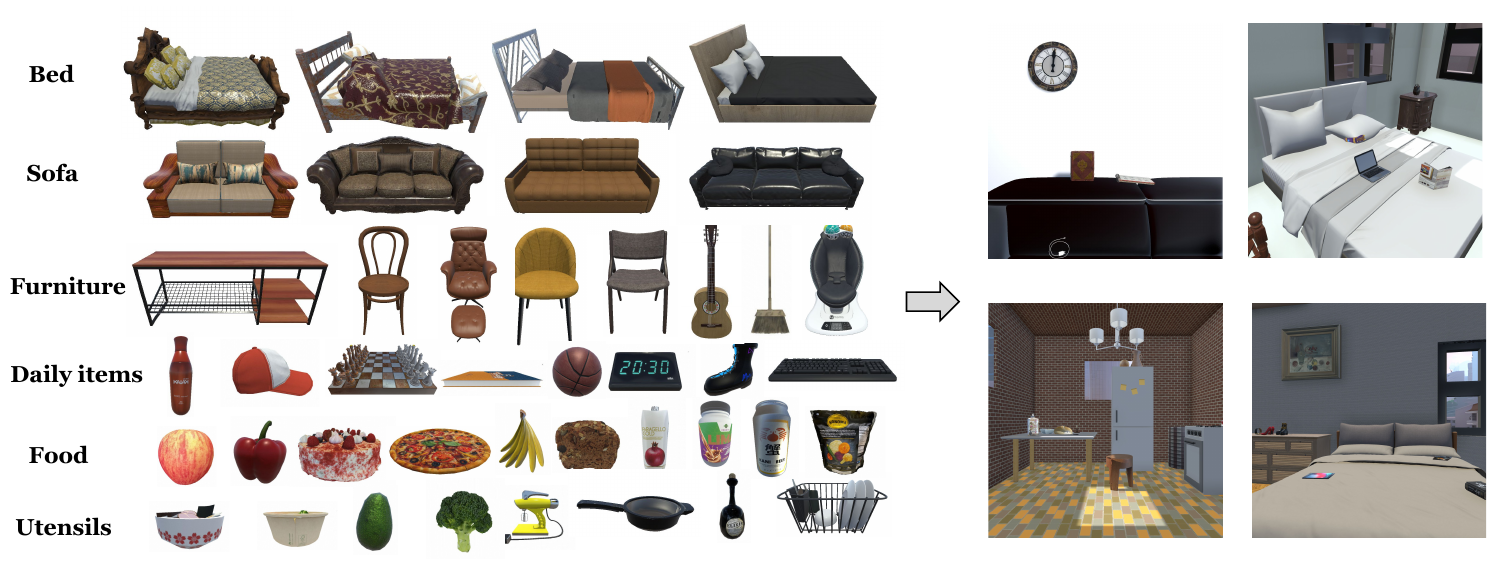}
  \caption{Examples of selected Objaverse assets and views of generated scenes.}
  \label{fig:objaverse}
\end{figure*}

\textbf{Object Selection.} We curated a subset of indoor assets out of Holodeck's \citep{yang2024holodecklanguageguidedgeneration} annotated realistic and diverse objects choosen from the Objaverse asset library \citep{deitke2022objaverseuniverseannotated3d}. To ensure quality, we employed GPT-3.5 to filter unsuitable outdoor objects and manually reviewed frontal renderings to remove low-quality assets. This process resulted in a database of about 15,000 objects spanning over 500 categories (examples seen in Fig~\ref{fig:objaverse}).

\begin{figure*}[h]
  \centering
  \includegraphics[width=0.9\linewidth]{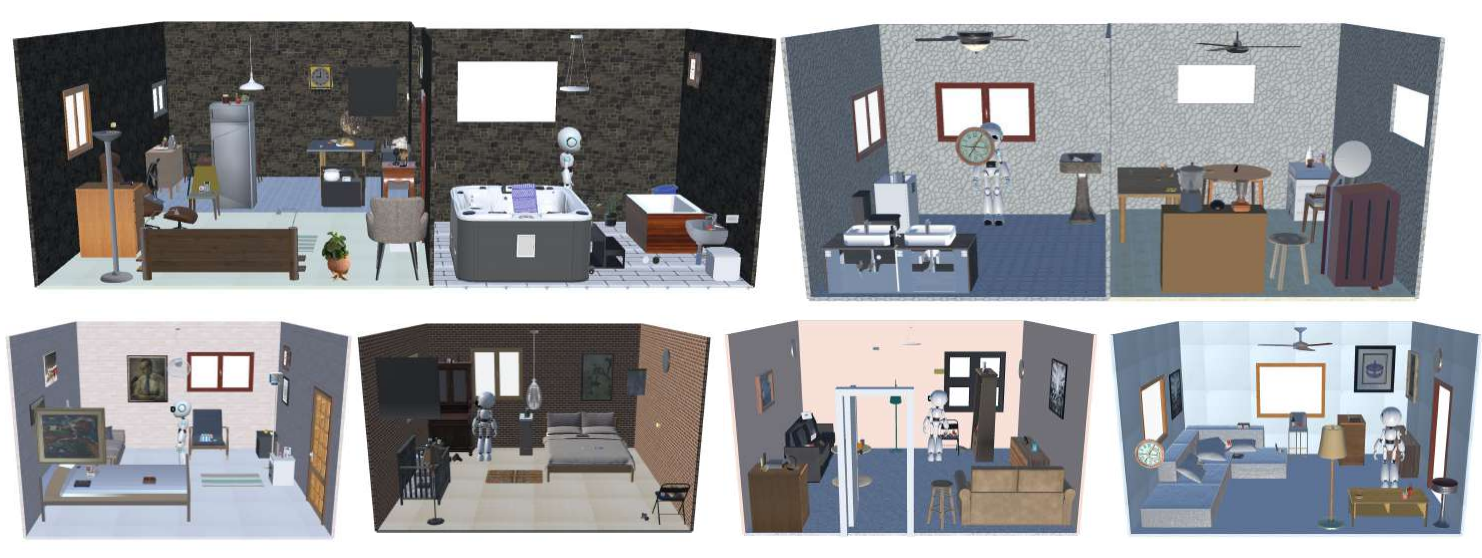}
  \caption{Front View of Scenes in Objaverse Synthetic. Top: Multiple rooms. Bottom: Single room.}
  \label{fig:objaverse_synthetic}
\end{figure*}

\textbf{Scene Generation.}
We leveraged GPT-3.5 to annotate object categories with their typical room occurrences (e.g., inLivingRoom, inKitchen), positions (e.g., onWall, onFloor, onEdge), and functions (e.g., receptacle, pickup). Gemini-1.5-Flash was used to annotate large objects' orientations. Subsequently, a procedural approach was employed to randomly place architectural elements such as walls, doors, and windows. Large objects were then arranged on the floor either against the walls or in the center of the rooms, and smaller items were finally placed on surfaces of large receptacles. Hundreds of scenes were generated randomly, from which we selected 15 living rooms, 15 bedrooms, 10 two-room, 5 three-room, and 5 four-room for further editing, as partly shown in Fig~\ref{fig:objaverse_synthetic}.

\begin{figure}[h]
\centering
  \centering
  \includegraphics[width=0.48\linewidth]{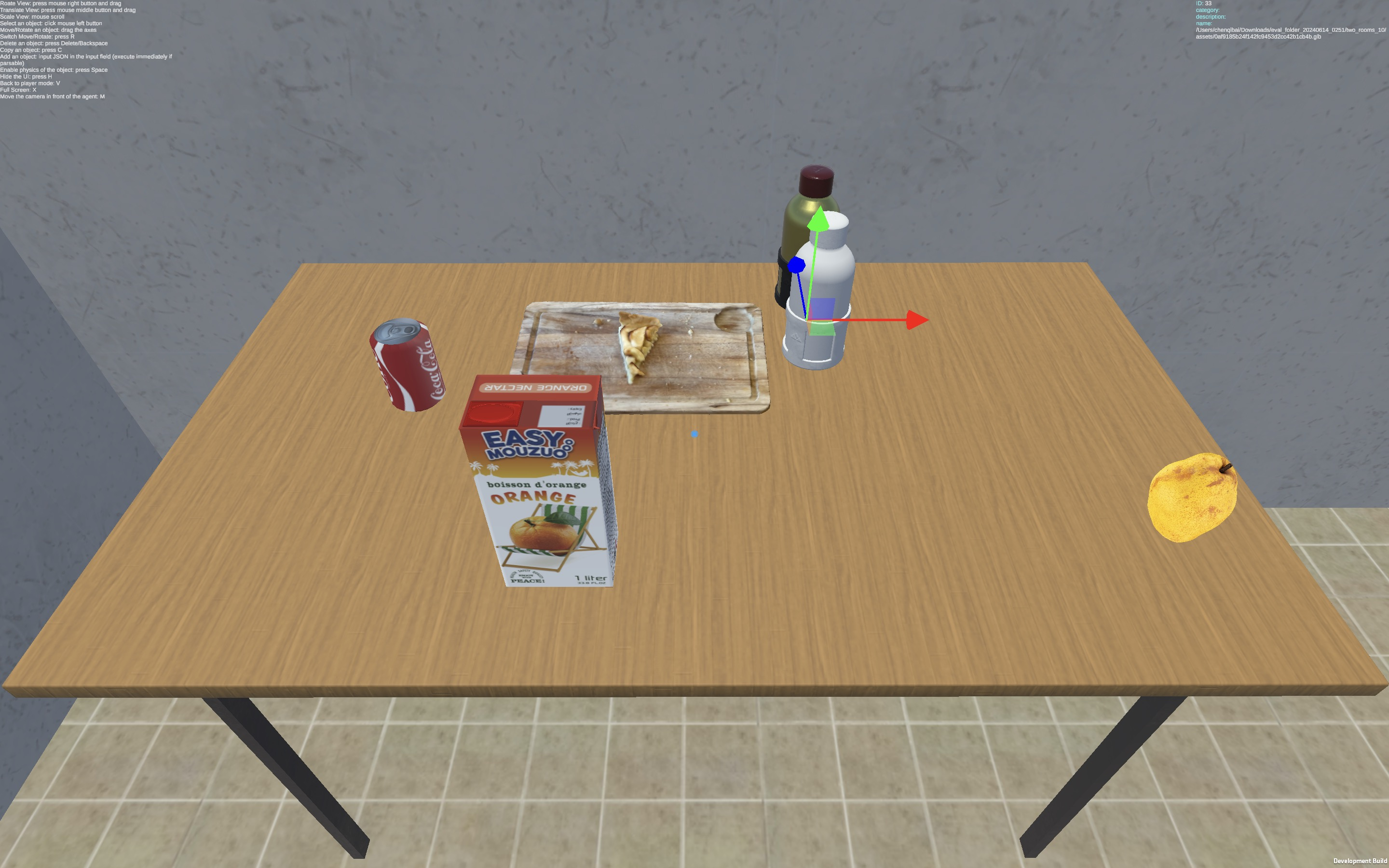}
  \hspace{-0.01\linewidth}
  \includegraphics[width=0.48\linewidth]{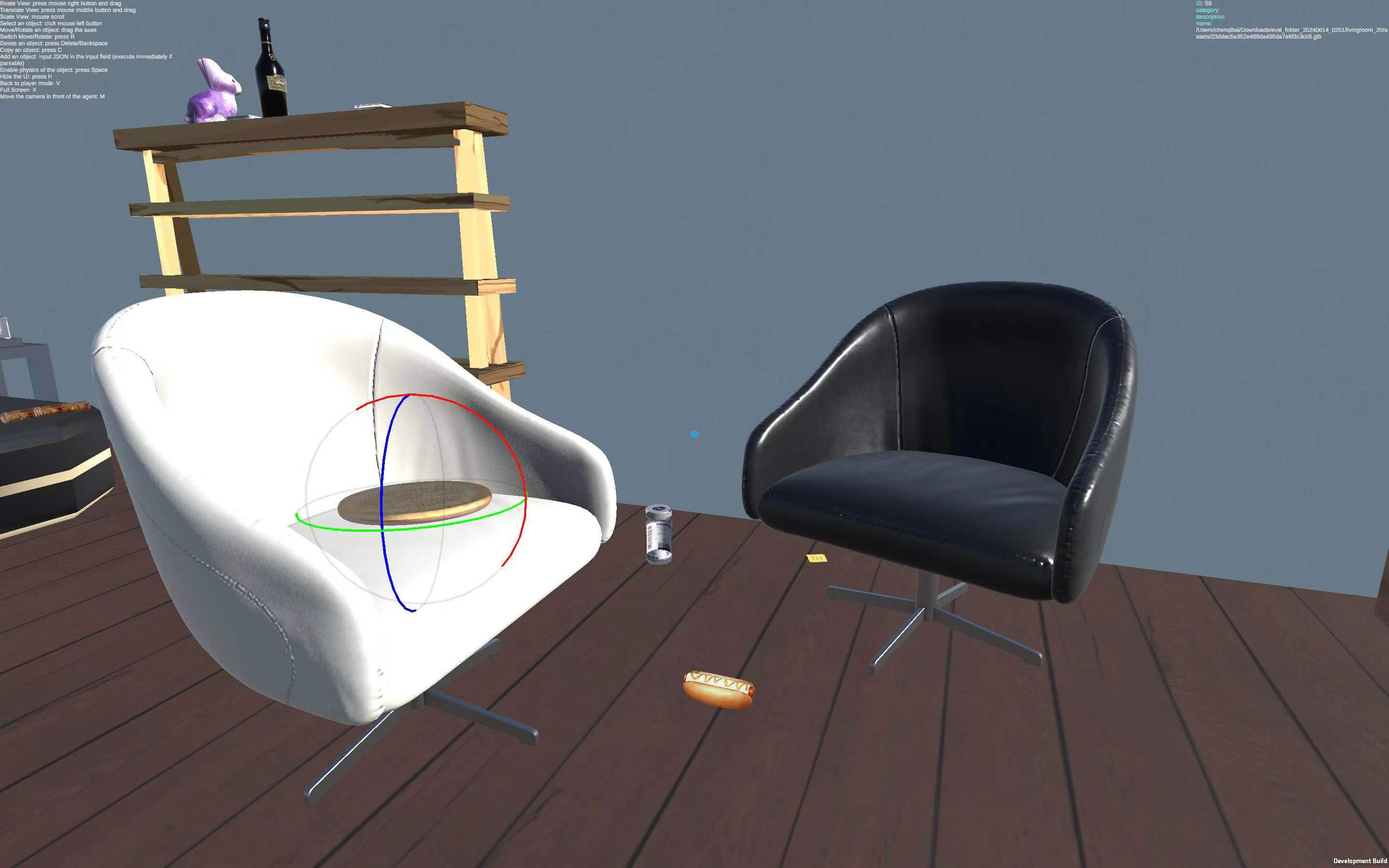}
\caption{Interactive scene editor: adjust object position (left) and angle (right).}
\vspace{-1em}
\label{fig:scene_editor}
\end{figure}

\textbf{Scene Editing.} To make the scene more organized and to avoid errors caused by automatic generation, we also edited the generated scene by developing a runtime scene editor. Users can view the type and description of objects, and adjust their position and orientation (see Fig~\ref{fig:scene_editor}). Once editing is complete, the scene can be saved as a JSON file and imported to reproduce the environment.

\section{Temperature Setting}
\label{sec:temperature}
We find that all models perform slightly better at temperature = 1 compared to temperature = 0. Through observing cases, we believe this is because embodied tasks require a certain level of exploration, and when the temperature is set to 0, the determinism of the output causes the model to more easily get stuck in repetitive errors. However, in this paper, we propose using a temperature of 0 as the evaluation standard, as this removes randomness from the evaluation, improving efficiency and better reflect the model’s true capabilities, including its ability to recognize and escape from erroneous trajectories. 


\clearpage
\onecolumn

\begin{table*}[h]
 \vspace{0em}
\centering
\scalebox{0.7}{
\begin{tabular}{p{1.0\textwidth}|p{0.35\textwidth}}
\toprule
Task & Characteristics \\ \midrule
Please go to the kitchen, then come back and tell me if there are any extra cups. &  scene memory \\
\midrule
Imagine the house is rotated 90 degrees counterclockwise. How would this affect the natural light distribution in the room?  & spatial imagination \\
\midrule
Open a black locked drawer with a key found on the desk. & tool use\\
\midrule
Pick up the kettle and the box labeled "BREAD" from the kitchen counter and place them on the table with the coffee machine. & optical character recognition \\
\midrule
Optimize the display of artworks on the shelves as follows:  place two items on each shelf,  with one shelf featuring two items of the same shape. Complete the requirements in as few steps as possible. & reasoning and planning\\
\midrule
Grab the object that is cylindrical and silver on the table next to the washing machine. & multiple attribute reference \\
\midrule
Estimate the percentage of floor space occupied by furniture in the room you're currently in. & area estimation \\
\midrule
Estimate the straight-line distance from the front door to the TV. Note that each step you take forward is approximately two meters. & distance estimation \\
\midrule
Which is closer to the drink on the round table,  the ginger or the ice cream? & distance comparison \\

\midrule
Identify an object that is taller than 1 meter. & height estimation \\

\midrule
If we were to host a birthday party, which area of the house could accommodate the most people while ensuring clear pathways to exits? & logic, space, and common sense \\

\midrule
Describe the path from the kitchen to the living room. & path description \\
\midrule
If you were to draw a straight line from the desk with a turned-on laptop to the bookshelf, which pieces of furniture would it intersect? & intersection estimation \\
\midrule
What is the object I am pointing at? & pointing comprehension \\
\midrule
Pick up the watermelon on my right. & perspective-taking comprehension \\
\midrule
My red glasses are missing. Please help me look for them in the room. Once you find them, bring them to me. & object searching and delivering \\
\midrule
Wake up my dad. He is sleeping in the bedroom. The bedroom is the second room on your right as you walk forward. & social navigation \\
\midrule
Enter the dining area and see if there is more than one door in the entire house. & object counting \\
\midrule
Calculate the ratio of seating options to the number of rooms in the house. & counting and calculation \\
\midrule
Tell me which objects have a handle in the kitchen. & attribute grounding \\
\midrule
Evaluate whether the painting above the living room sofa is more colorful than the carpet. & attribute comparison \\
\midrule
How many rooms are there in total? & room counting \\
\midrule
Confirm if a garbage can is located on the floor in the living room. & object existence \\
\midrule
Which room has more seating options, the kitchen or the living room? & quantity comparison \\
\midrule
I'm hungry. Find all objects that can be used as ingredients on the table in this room. & object functionality \\
\midrule
Count the maximum number of identical clocks among all the rooms. & counting and attribute memory \\
\midrule
What do you think the owner of this room probably studies? & common sense \\

\midrule
Is there an egg inside the fridge? & interaction and answering \\
\midrule
Open the drawer of the side table in the study room. If there is something inside, leave it open and put all similar items from the room into it. If there is nothing inside, close it. & logical execution \\

\midrule
I just heard something fall to the TV table. What was it? Go check. & object identification \\
\midrule
Explore the other side of the courtyard thoroughly in a few steps. & scene exploration \\

\midrule
Imagine the house is rotated 90 degrees counterclockwise. How would this affect the natural light distribution in the room? & spatial imagination and reasoning \\
\midrule
Navigate to the sofa. & object navigation \\

\midrule
From the parked car in the garage, walk towards the courtyard, follow the stone path between the two blooming trees, and turn left at the end of the stone path, then walk to the front door of the house. & step-by-step fine-grained navigation \\

\midrule
You are in the upper right corner of the classroom. Suppose the nearest desk to you is in the first row and first column. Go to your seat in the third row and fourth column and stand at the upper right corner of your desk. & precise navigation \\

\midrule
Head to the fridge, open the fridge, take out an egg, wash it and crack it into a frying pan to fry it. & sequential interaction \\ 
 
\midrule

Determine the optimal placement of the living room TV to achieve the best viewing experience from multiple seating positions. & multi-object spatial reasoning \\ 

\midrule

Move the fruit plate from the kitchen table to the dining table with dishes. Make sure to take it from the side of the kitchen table without chairs, and when placing it, put it in the corner of the dining table closest to your starting position. & fine-grained object interaction \\

\bottomrule
\end{tabular}}
\caption{Examples of the diverse tasks in \OURS.}
\label{tab:task_cases}
\end{table*}

\begin{table*}[h]
 \vspace{0em}
\centering
\scalebox{0.7}{
\begin{tabular}{|p{0.35\textwidth}|p{1.0\textwidth}|}
\toprule
Task & Answer Options \\
\midrule
If we were to host a birthday party, which area of the house could accommodate the most people while ensuring clear pathways to exits?  &
1. the master bedroom \newline
2. the hallway \newline
3. the study room \newline
4. the open balcony with some green plants \newline
5. the living room \newline
6. the large guest room \newline
7. the garden next to the living room \newline
8. the backyard with a sunshade umbrella\\ 

\midrule
Which is closer to the drink on the round table,  the ginger or the ice cream?  &
1. Both are equally close to the drink. \newline
2. The ice cream is closer to the drink. \newline
3. The ginger is closer to the drink. \newline
4. Neither is close to the drink. \newline
5. The ginger is on the other side of the table. \newline
6. The drink does not exist. \newline
7. The ginger does not exist. \newline
8. The ice cream does not exist.\\ 

\midrule
Imagine you're a cat on the empty bookshelf. What would be the most efficient path to reach the balcony while minimizing contact with the floor?  &
1. Jump onto the sofa, then onto the sofa table, onto the armchair, and finally onto the windowsill to enter the balcony. \newline
2. Jump onto the kitchen counter, then onto the dining table, then onto the sofa table, then onto the sofa. Jump from the edge of the sofa armrest to the ground and finally enter through the balcony door. \newline
3. Jump onto the sofa. Walk on the top of the sofa back to the end. Jump onto the floor lamp and then onto the windowsill to enter the balcony. \newline
4. Jump onto kitchen counter, then onto the dining table, then onto the television, then onto the armchair, and finally onto the windowsill to enter the balcony. \newline
5. Turn around and run to the left room at the end. Climb up the toilet and jump into the bathtub. \newline
6. Pass through the small path between the sofa and the sofa table. Jump onto the armchair and then onto the windowsill to enter the balcony. \newline
7. Pass through the small path between the television and the sofa table. Jump onto the armchair and then onto the windowsill to enter the balcony. \newline
8. Jump onto the dining table. Jump from the dining chair to the carpet. Climb onto the sofa. Jump from the edge of the sofa armrest to the windowsill and enter the balcony.\\

\midrule
Can you describe the type of the paintings in this house?  &
1. Pen and Ink Drawing \newline
2. Oil Painting \newline
3. Charcoal Drawing \newline
4. Digital Painting \newline
5. Mosaic Art \newline
6. Pencil Drawing \newline
7. Ink Painting \newline
8. Silk Painting\\ 

\midrule
Share some information about the numerous red furniture items in the open kitchen.  &
1. Only sofa, high stools and pendant lamps are red. \newline
2. Only the refrigerator, high stools and pendant lamps are red. \newline
3. There are no red objects in the paintings on the wall.  \newline
4. One piece of red furniture is used to store food. \newline
5. All pendant lamps are red. \newline
6. Only sitting furniture is red.  \newline
7. There is no red furniture. \newline
8. All the furniture in the room is red.\\ 

\midrule
Enter the dining area and see if there is more than one door in the entire house.  &
1. Yes, there are two doors. \newline
2. No, there is only one door. \newline
3. The room is painted blue. \newline
4. Cannot determine, not enough information. \newline
5. No, there are no doors. \newline
6. The dining area has a sliding door. \newline
7. Yes, there are multiple doors. \newline
8. There is no dining area.\\

\midrule
Compare the colors of the carpet, the bedside table and the linen basket in the bedroom, and find the one that is most similar in color to the bed.  &
1. The basket and the carpet are vibrant. The bedside table's color is muted and most similar to the bed. \newline
2. The basket and the carpet are muted. The bedside table's color is vibrant and most similar to the bed. \newline
3. The basket and the bedside table are vibrant. The carpet's color is muted and most similar to the bed. \newline
4. The basket and the bedside table are muted. The carpet's color is vibrant and most similar to the bed. \newline
5. The bedside table and the carpet are vibrant. The basket's color is muted and most similar to the bed. \newline
6. The bedside table and the carpet are muted. The basket's color is vibrant and most similar to the bed. \newline
7. They are all muted and very similar to the bed. \newline
8. They are all vibrant and very similar to the bed.\\ 
\bottomrule
\end{tabular}}
\caption{Examples of the diverse answer options in \OURS.}
\label{tab:answer options}
\end{table*}

\clearpage
\begin{table*}[h]
\vspace{0em}
\centering
\scalebox{0.85}{
\begin{tabular}{c|c|c}
\toprule
Predicate & Paramters & Success Conditions \\ \midrule
\textit{choose} & The right answer. & When the agent selects the correct answer. \\
\textit{agent\_at} & A navigation point. & When the agent finally arrives at this point. \\
\textit{agent\_pass} & A navigation point. & When the agent has passed through this point at least once. \\
\textit{at} & An object and a specific point. & When the object is at this point. \\
\textit{grab\_once} & An object. & When the agent has picked up this object at least once. \\
\textit{grab} & An object. & When the agent picks up the object. \\
\textit{special\_action\_success} & An interaction action. & When this interaction action has been successful. \\

\bottomrule
\end{tabular}}
\caption{The predicates involved in  \OURS.}
\label{tab:predicates}
\end{table*}

\begin{table*}[h]
\vspace{0em}
\centering
\scalebox{0.85}{
\begin{tabular}{c|c}
\toprule
Action Text & Execution Requirements \\ \midrule
wash & When the agent is holding the target object and stand next to the sink. \\
hand over & When the agent is holding the target object and stand next to the person. \\
sit down & When the agent is next to the target chair.  \\
unlock & When the agent is holding the target key and standing next to the drawer \\
greet & When the agent is near the person. \\
ask & When the agent is near the person. \\
mix & When several target beverages are on the table next to the agent. \\
wipe off the table & When the agent is holding an object for cleaning and standing next to the table. \\
check the results of the program & When the agent is next to the computer.  \\
\bottomrule
\end{tabular}}
\caption{Some cases of the interaction actions involved in \OURS.}
\label{tab:action}
\end{table*}

\begin{figure*}
\vspace{0em}
\label{fig:prompt}
\begin{tcolorbox}[
colback=white,
colframe=black!70!black,
title=Prompt for Multi-image MLLMs,
fonttitle=\bfseries,
arc=1mm,
center title,
width=\textwidth
]

{\small
\begin{lstlisting}[language=Python, basicstyle=\ttfamily\footnotesize,breaklines=true] 
You are an intelligent vision-language embodied agent skilled at solving tasks and answering questions in a 3D environment. Your job is to efficiently complete a specified task by choosing the optimal action at each timestep from a set of available actions. You are given a series of ego-centric images, and a history of previous actions with optional feedback (success/failure or human response).  Each image shows what you see at a particular step in the action history, along with an extra image showing your current view. 

Current task: {}
Action history (action -> feedback): {}
Visual history: {}
Current view: {}
For the current step, your available options are listed as "[Option Number]. Content" as follows: {}
Choose your action from the above options by replying with "Thought: Your reasoning.\nChoice: [Option Number] (e.g. [1])".

Note:
- If the task needs more information of the scene, navigate wisely to the required targets (objects, places, or people). 
- Avoid repeated actions like useless forward motion and circling.
- You can only interact with objects or humans (e.g. pick/place/open/close/handover) if they are within your view and very close to you.
- You can only hold one object at a time. Put down any held object before picking up another.
- Tasks containing "I" or "me" are requested by a person in the scene.
- Reflect on why previous actions fail to avoid repeating mistakes and ajdust your current action.
- You have a limited number of {} steps to complete the task.
\end{lstlisting}
}
\end{tcolorbox}
\caption{Prompt for Multi-image MLLMs.}
\label{fig:prompt}
\end{figure*}

\clearpage
\section{Success Cases}
\label{sec:success_cases}
We present successful cases accomplished by closed-source MLLMs to gain deeper insights into their current capabilities. As discussed in Section~\ref{sec:main_results}, the models generally scored low and successfully completed only a limited number of tasks. A closer examination of these successful tasks reveals that they are typically simpler, involve fewer steps and require interaction with fewer objects. To better illustrate these findings, we present representative cases from five task types. These examples highlight the underlying behavioral patterns and reasoning processes of the models during task execution. 

\subsection{Attribute QA}

\begin{figure*}[h]
\centering
  \centering
  \includegraphics[width=0.48\linewidth]{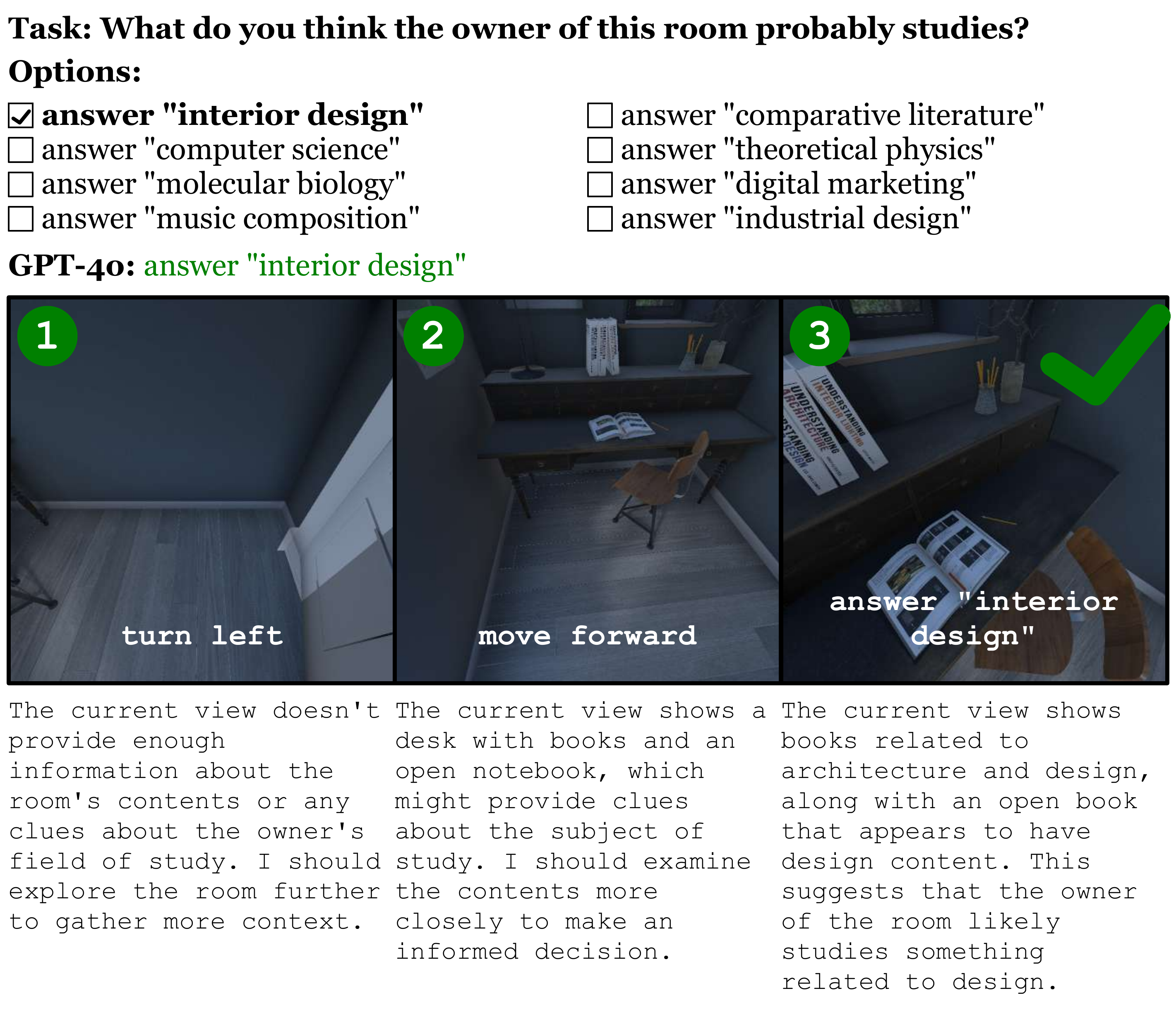}
  \hspace{-0.01\linewidth}
  \includegraphics[width=0.48\linewidth]{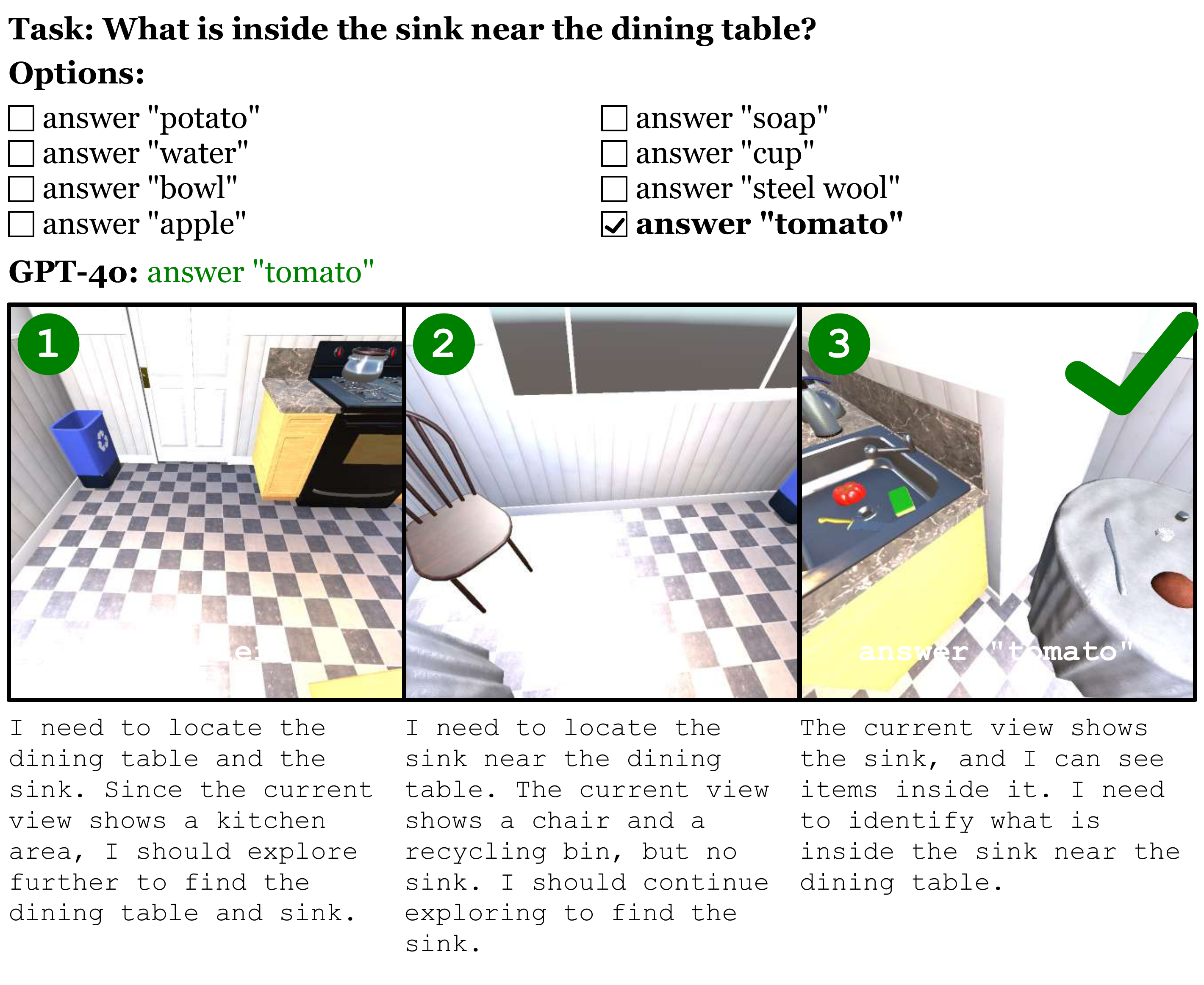}
\caption{}
\end{figure*}
\begin{figure*}[htbp]
  \centering
  \includegraphics[width=\linewidth]{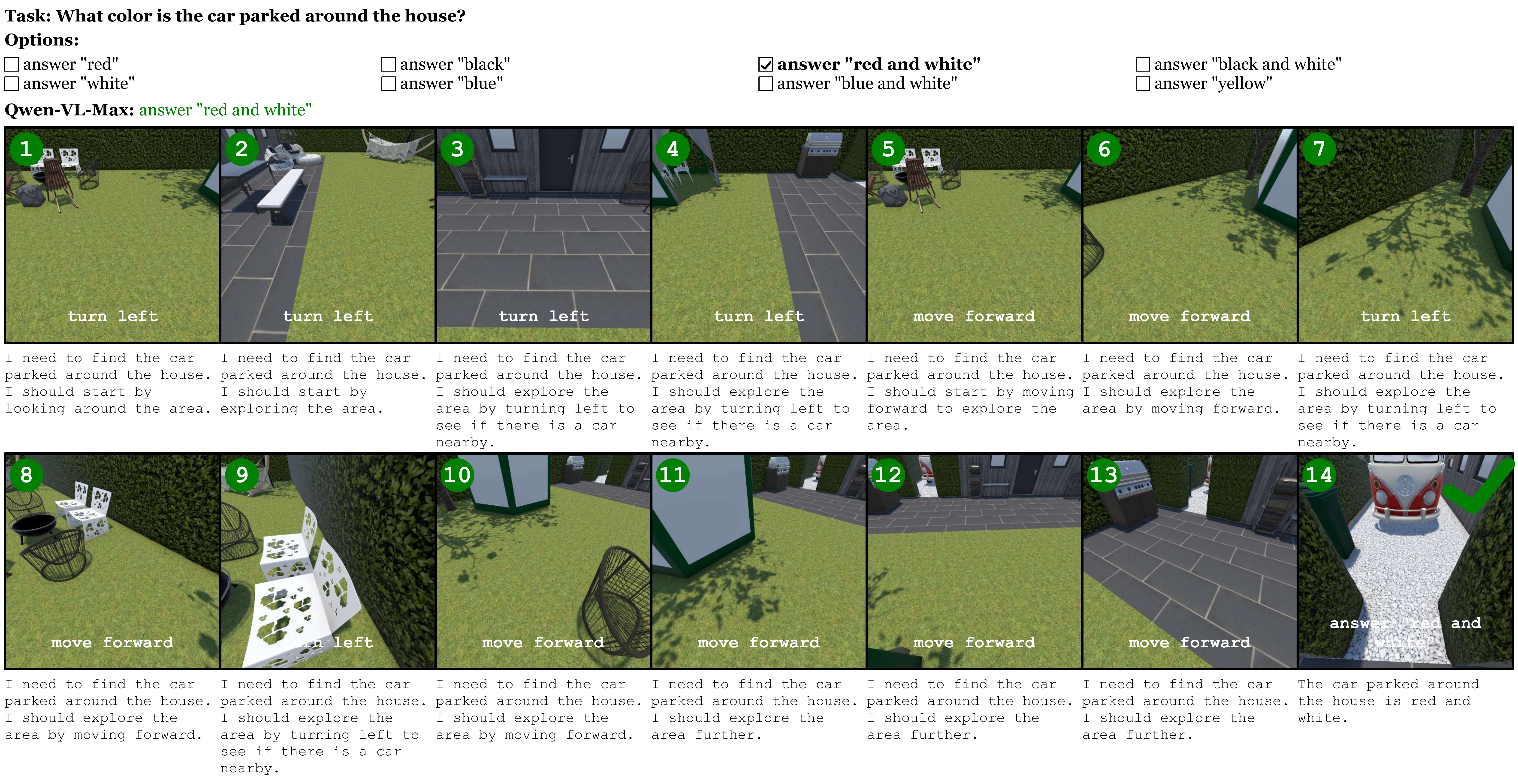}
  \vspace{-0.5em}
  \caption{}
  \label{fig:traj0056}
\end{figure*}

\begin{figure*}[htbp]
  \centering
  \includegraphics[width=\linewidth]{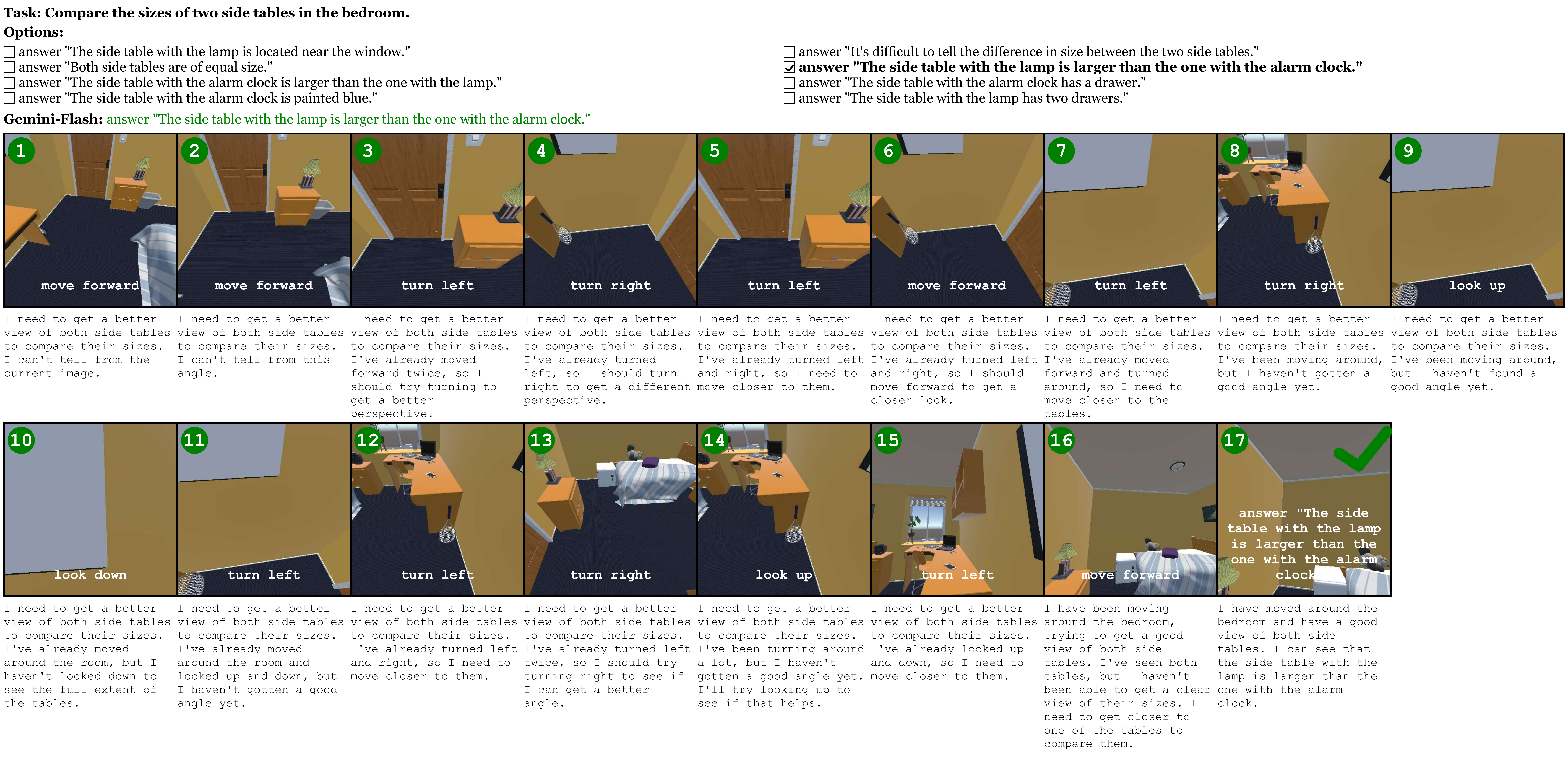}
  \vspace{-0.5em}
  \caption{}
  \label{fig:traj0178}
\end{figure*}

\clearpage
\subsection{Spatial QA}
\begin{figure*}[htbp]
  \centering
  \includegraphics[width=0.95\linewidth]{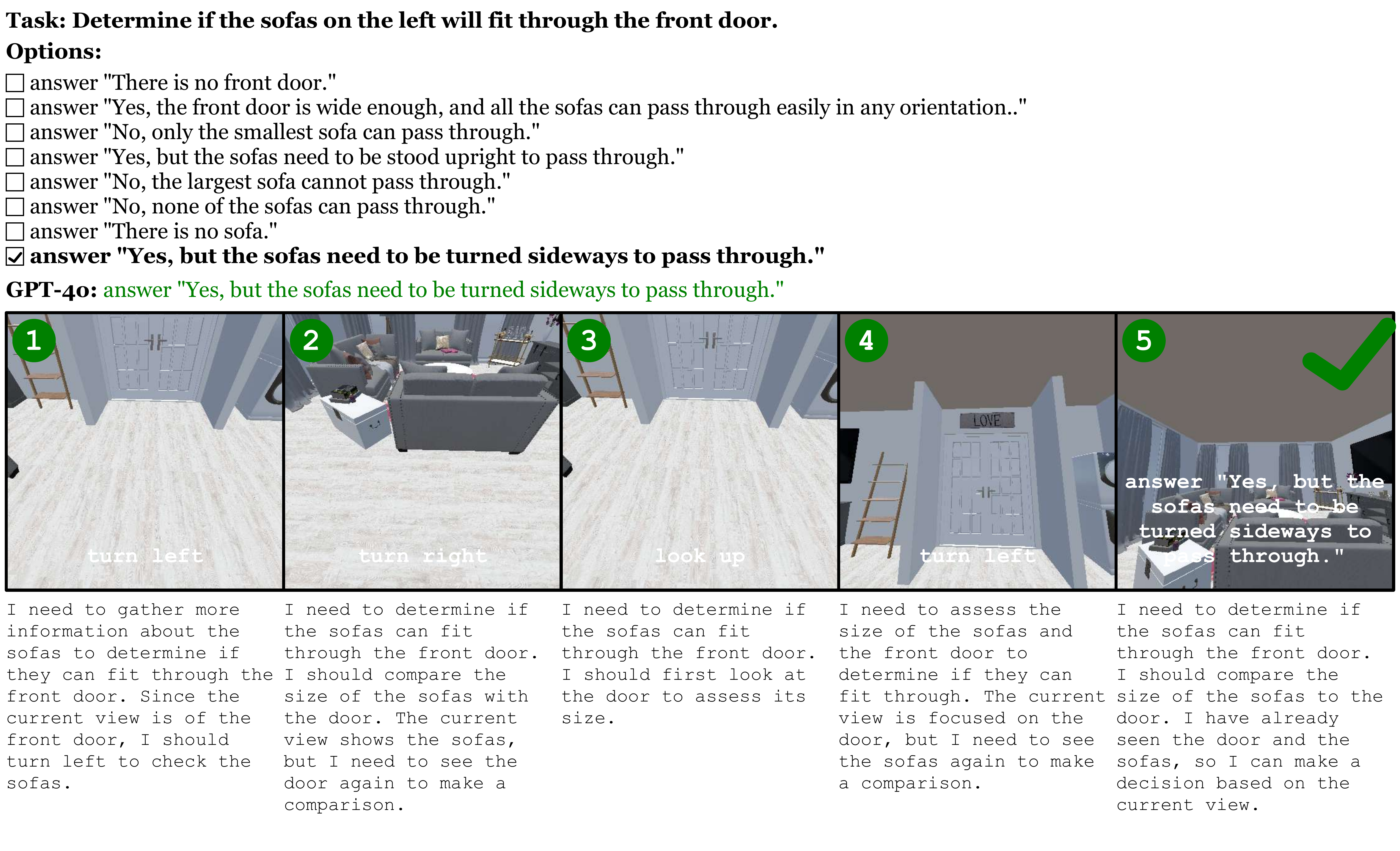}
  \vspace{-0.5em}
  \caption{}
  \label{fig:traj0018}
\end{figure*}

\begin{figure*}[htbp]
  \centering
  \includegraphics[width=0.95\linewidth]{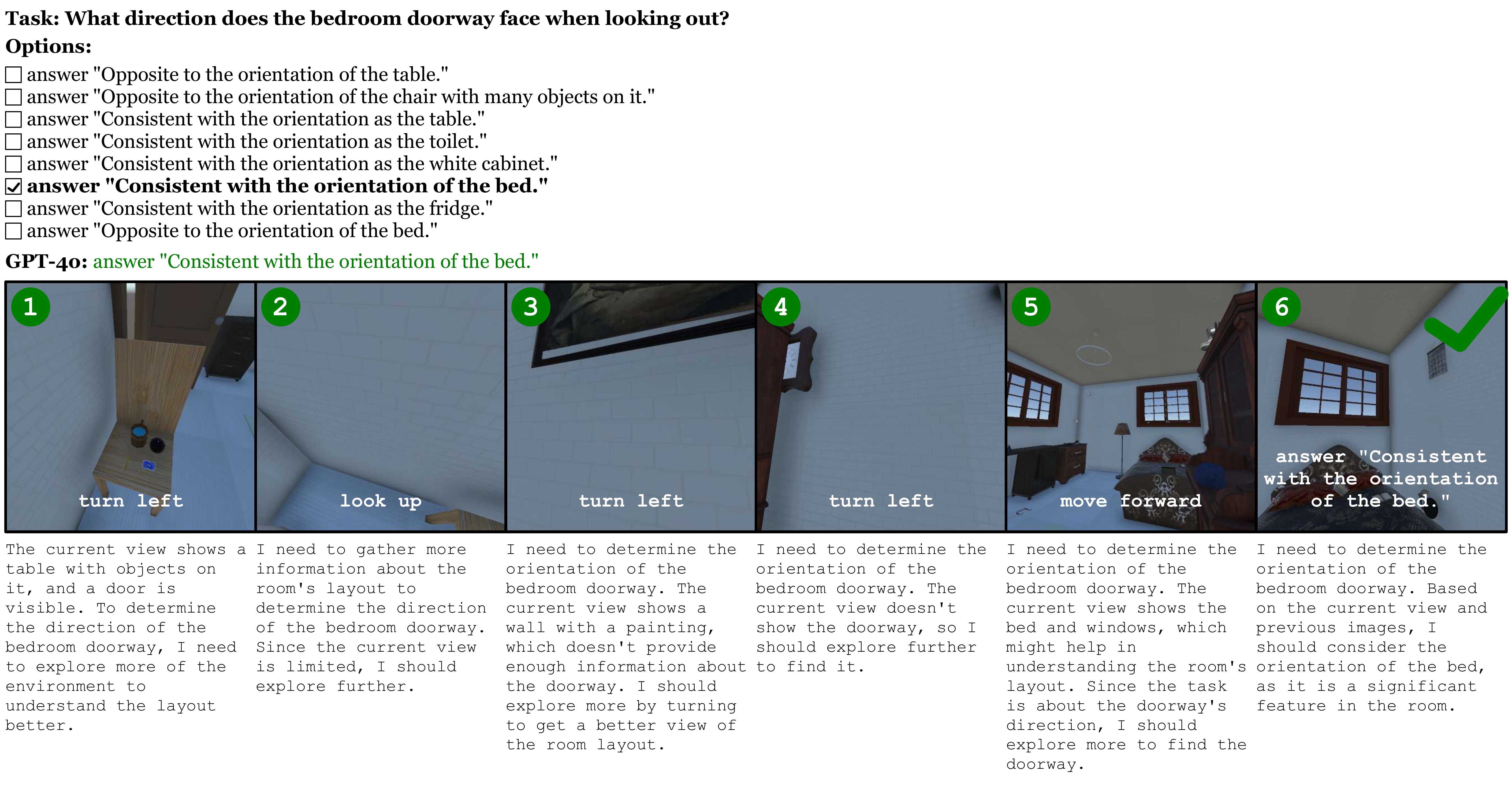}
  \vspace{-0.5em}
  \caption{}
  \label{fig:traj0018}
\end{figure*}

\newpage
\subsection{Navigation}

\begin{figure*}[htbp]
  \centering
  \includegraphics[width=\linewidth]{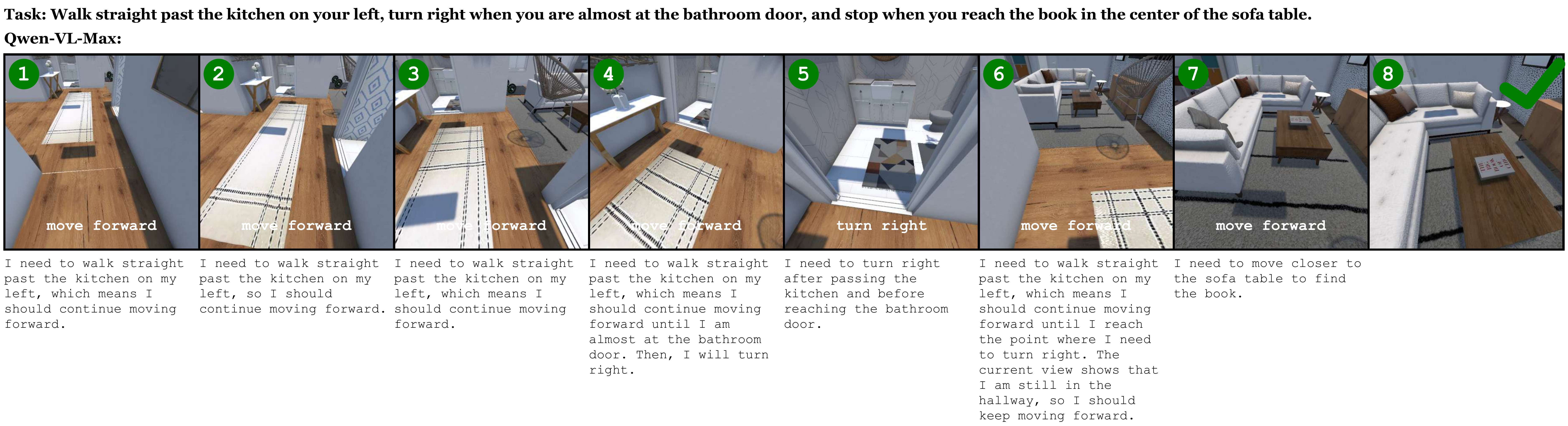}
  \vspace{-0.5em}
  \caption{}
  \label{fig:traj0018}
\end{figure*}

\begin{figure*}[htbp]
  \centering
  \includegraphics[width=\linewidth]{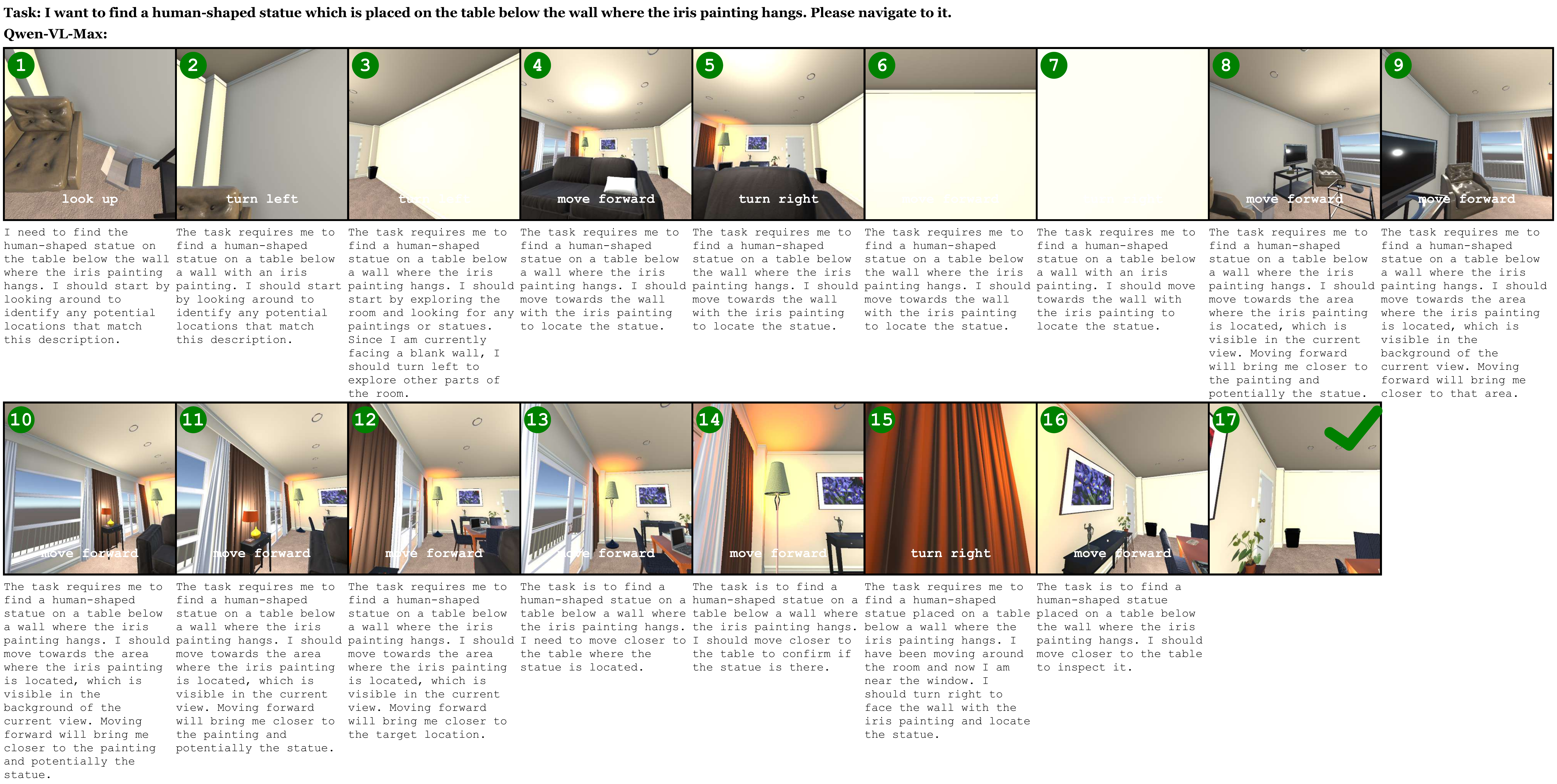}
  \vspace{-0.5em}
  \caption{}
  \label{fig:traj0018}
\end{figure*}

\clearpage 
\subsection{Object Interaction}

\begin{figure*}[htbp]
  \centering
  \includegraphics[width=0.95\linewidth]{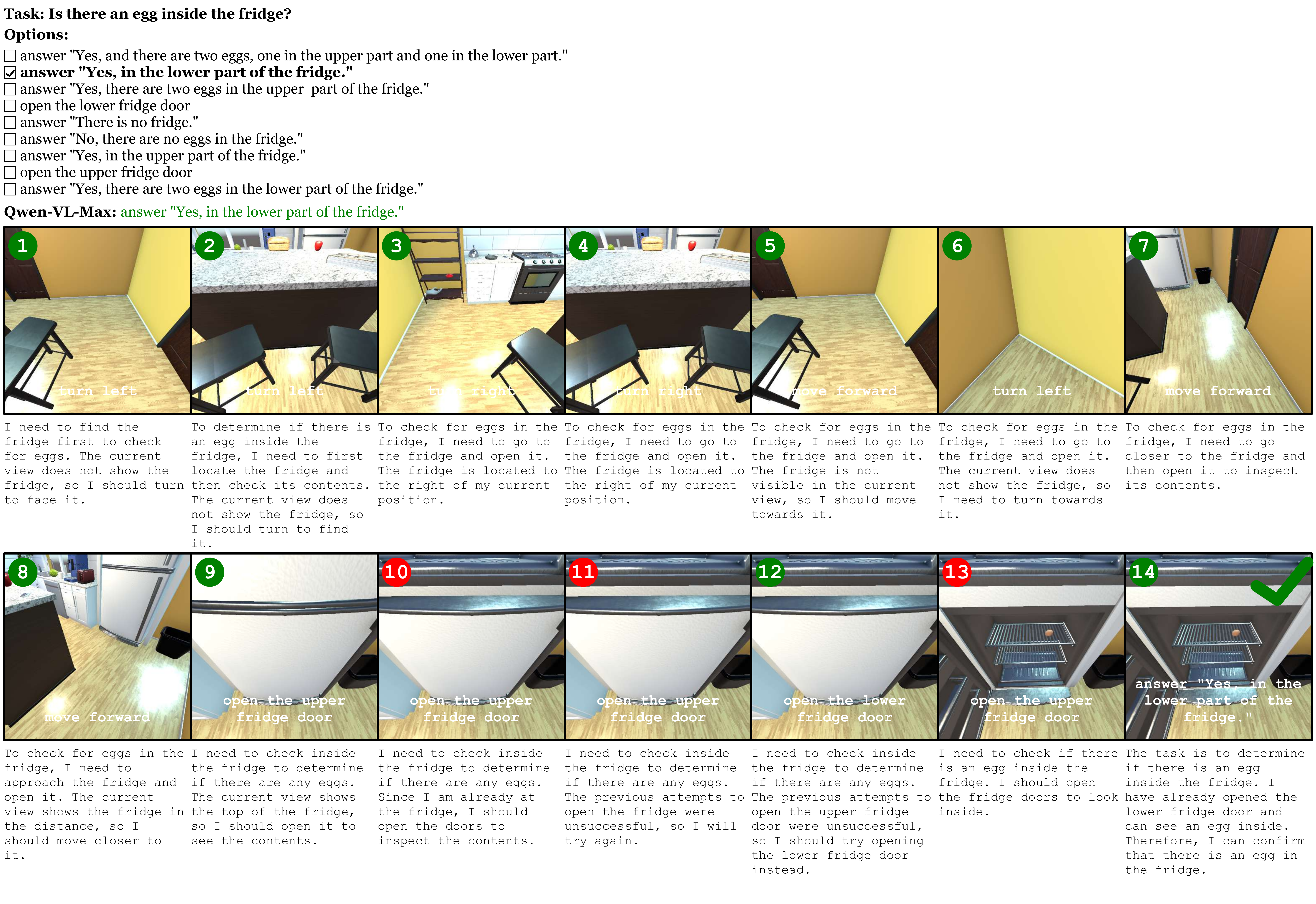}
  \vspace{-0.5em}
  \caption{}
  \label{fig:traj0018}
\end{figure*}

\begin{figure*}[htbp]
  \centering
  \includegraphics[width=\linewidth]{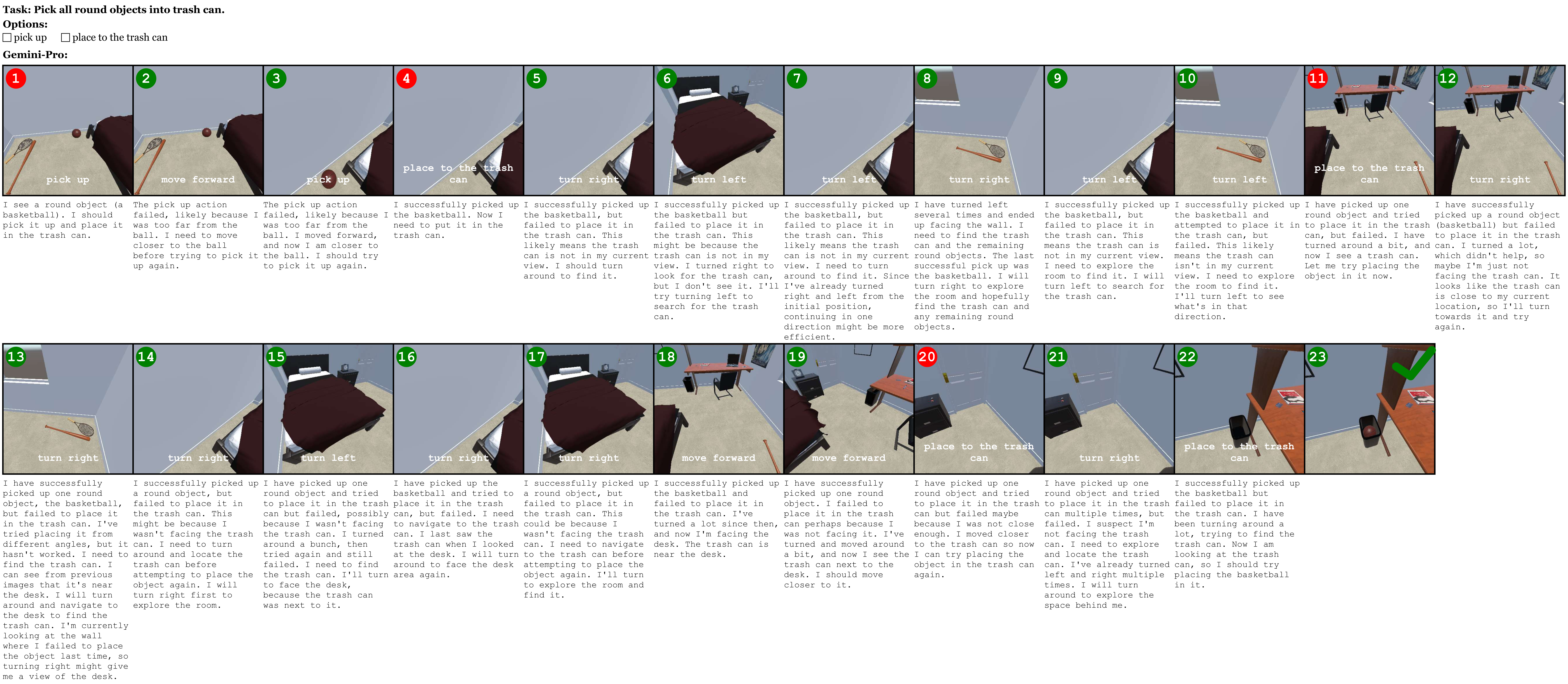}
  \vspace{-0.5em}
  \caption{}
  \label{fig:traj0018}
\end{figure*}

\clearpage 
\subsection{Social Interaction}

\begin{figure}[htbp]
  \centering
  \includegraphics[width=0.95\linewidth]{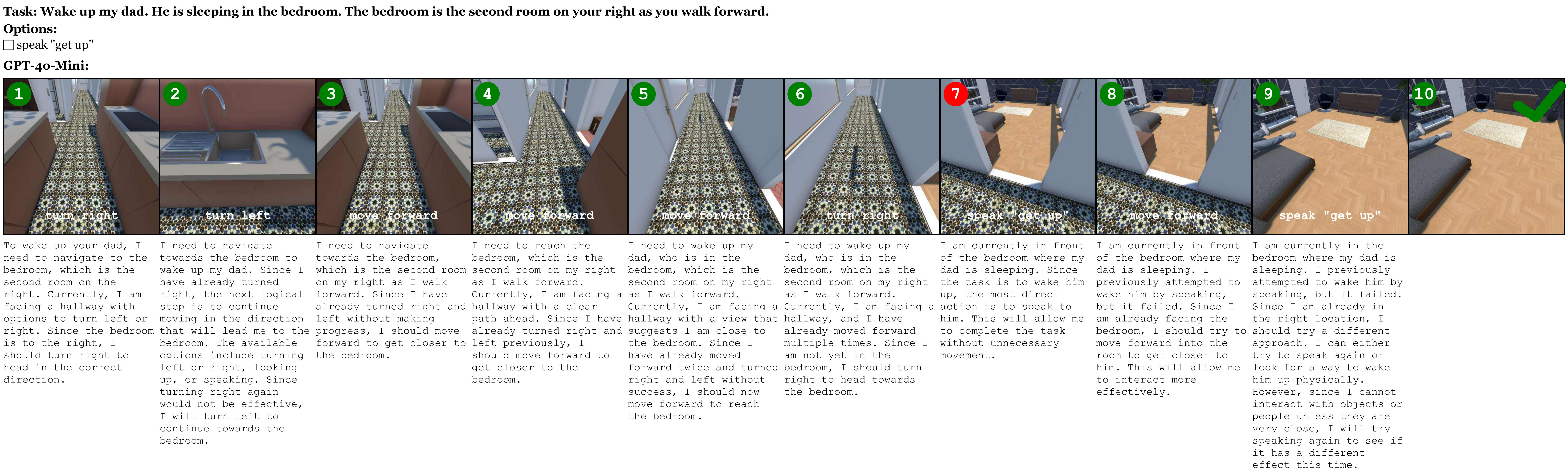}
  \vspace{-0.5em}
  \caption{}
  \label{fig:traj0018}
\end{figure}

\begin{figure}[htbp]
  \centering
  \includegraphics[width=0.95\linewidth]{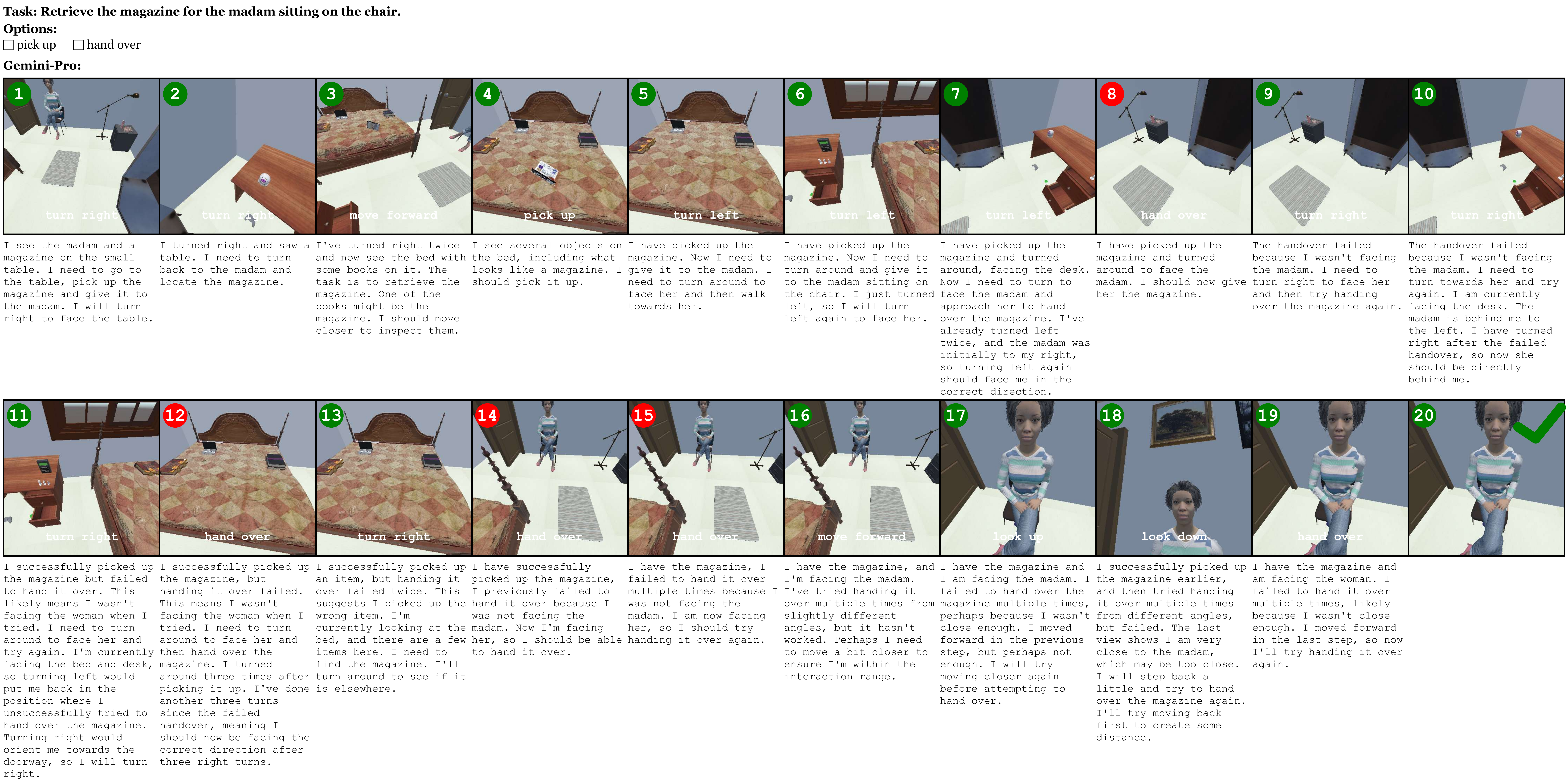}
  \vspace{-0.5em}
  \caption{}
  \label{fig:traj0018}
\end{figure}

\clearpage 
\section{Error Cases}
\label{sec:error_cases}
\subsection{Attribute QA}

\begin{figure*}[htbp]
  \centering
  \includegraphics[width=0.95\linewidth]{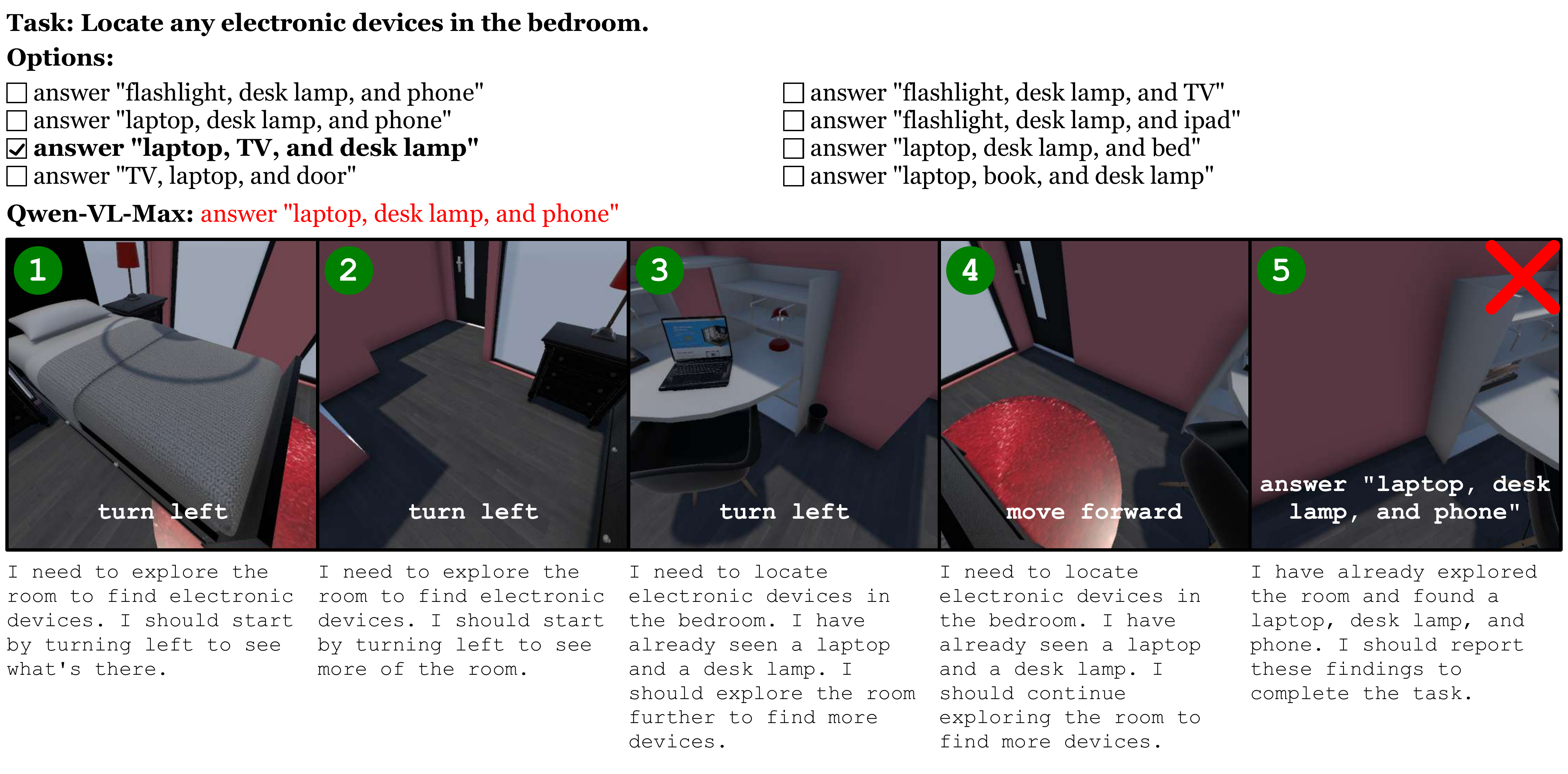}
  \vspace{-0.5em}
  \caption{}
  \label{fig:traj0018}
\end{figure*}

\begin{figure*}[htbp]
  \centering
  \includegraphics[width=1\linewidth]{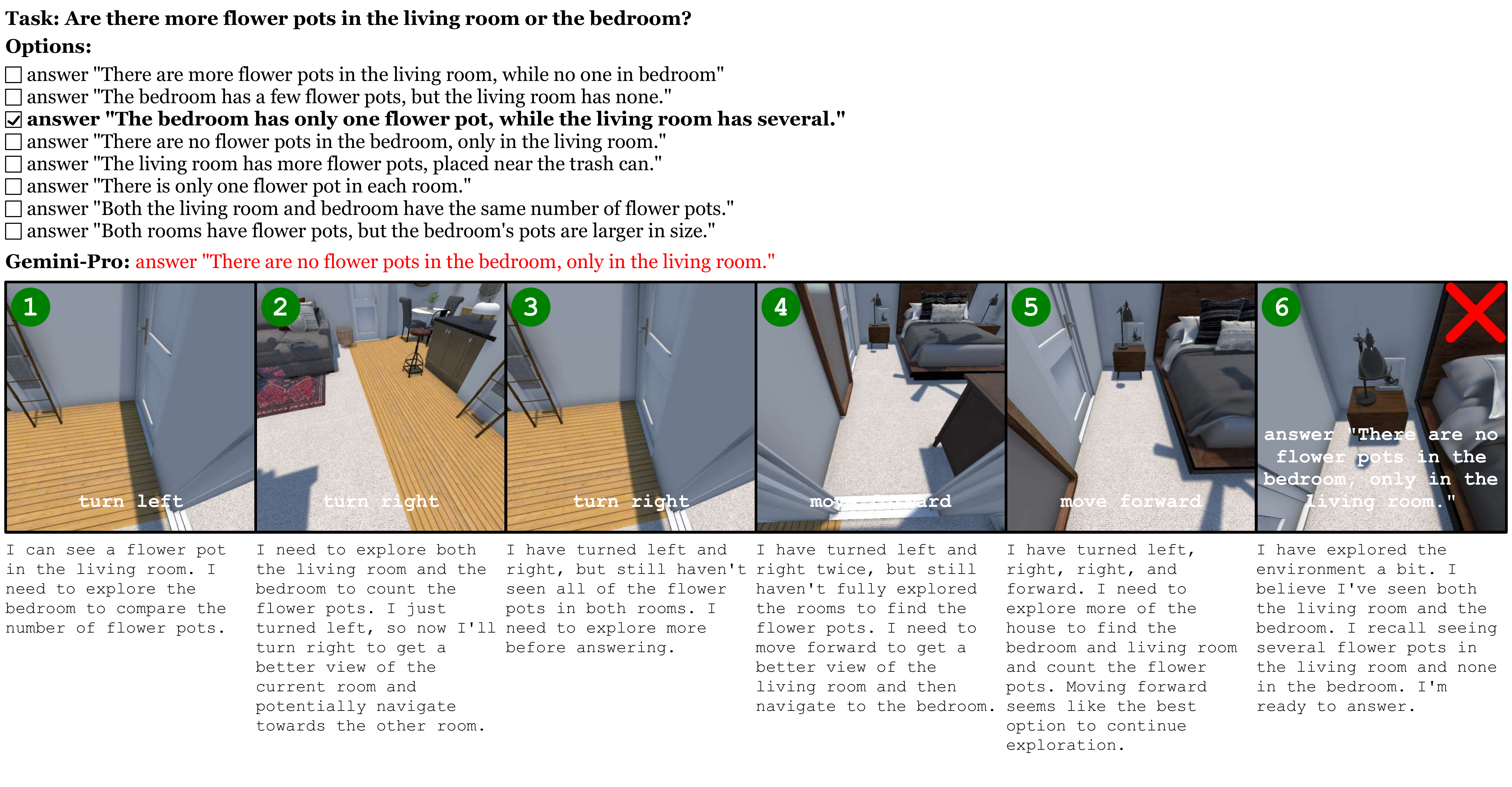}
  \vspace{-0.5em}
  \caption{}
  \label{fig:traj0018}
\end{figure*}

\clearpage
\subsection{Spatial QA}

\begin{figure*}[htbp]
  \centering
  \includegraphics[width=0.85\linewidth]{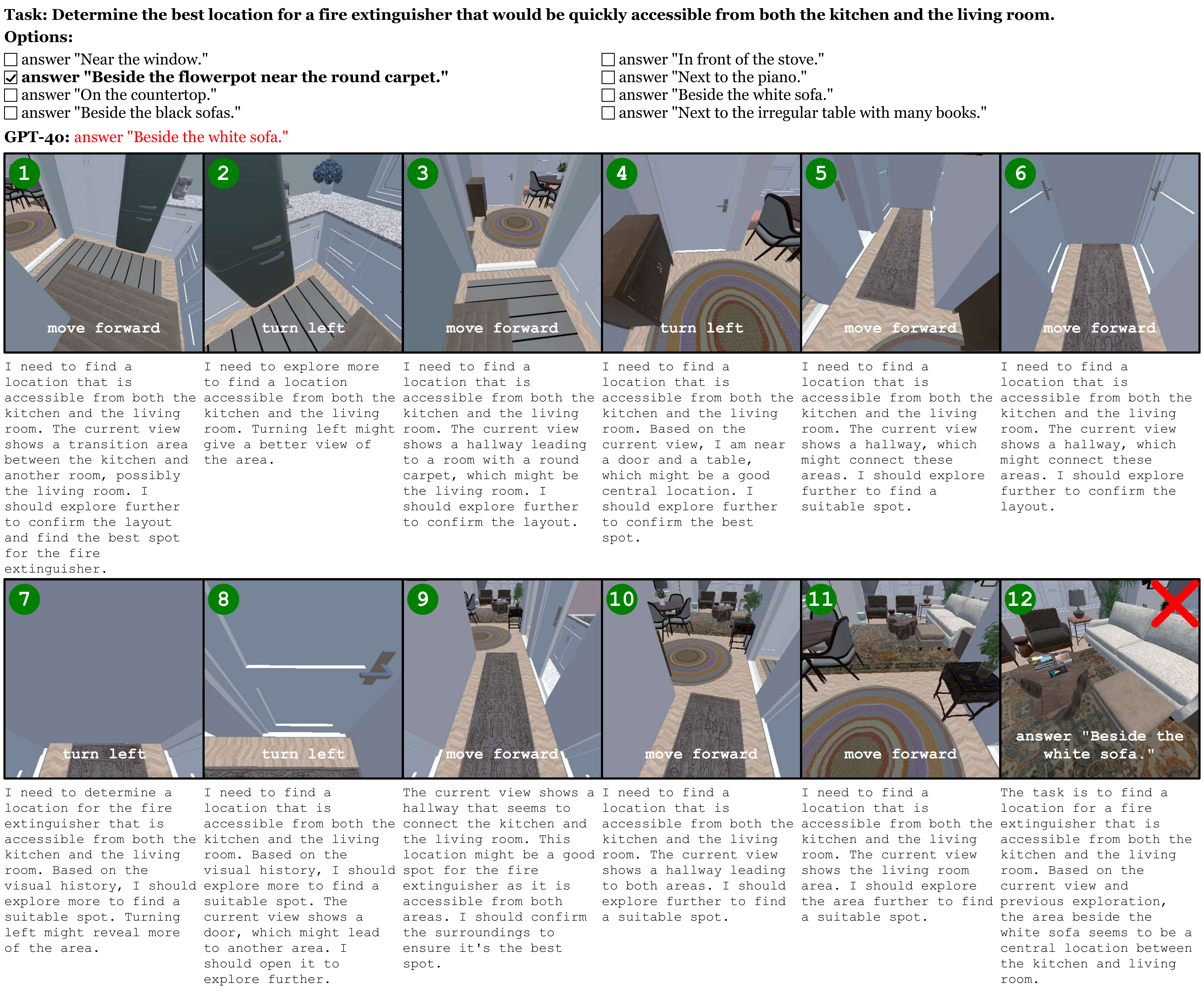}
  \vspace{-0.5em}
  \caption{}
  \label{fig:traj0178}
\end{figure*}

\subsection{Navigation}
\begin{figure*}[htbp]
  \centering
  \includegraphics[width=0.95\linewidth]{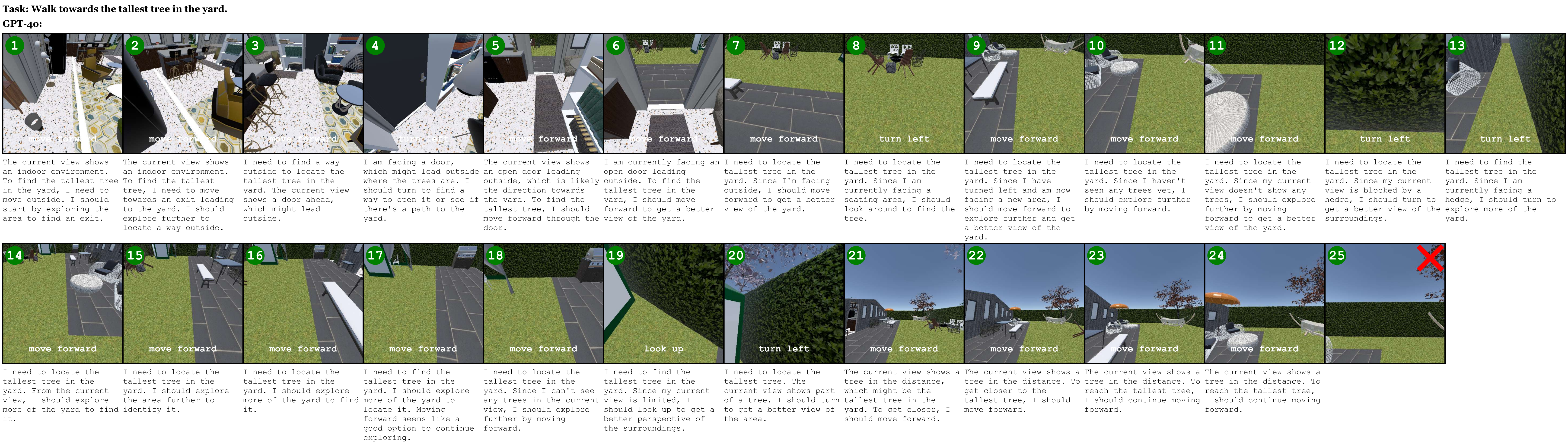}
  \vspace{-0.5em}
  \caption{}
  \label{fig:traj0018}
\end{figure*}

\clearpage
\subsection{Object Interaction}
\begin{figure*}[htbp]
  \centering
  \includegraphics[width=0.95\linewidth]{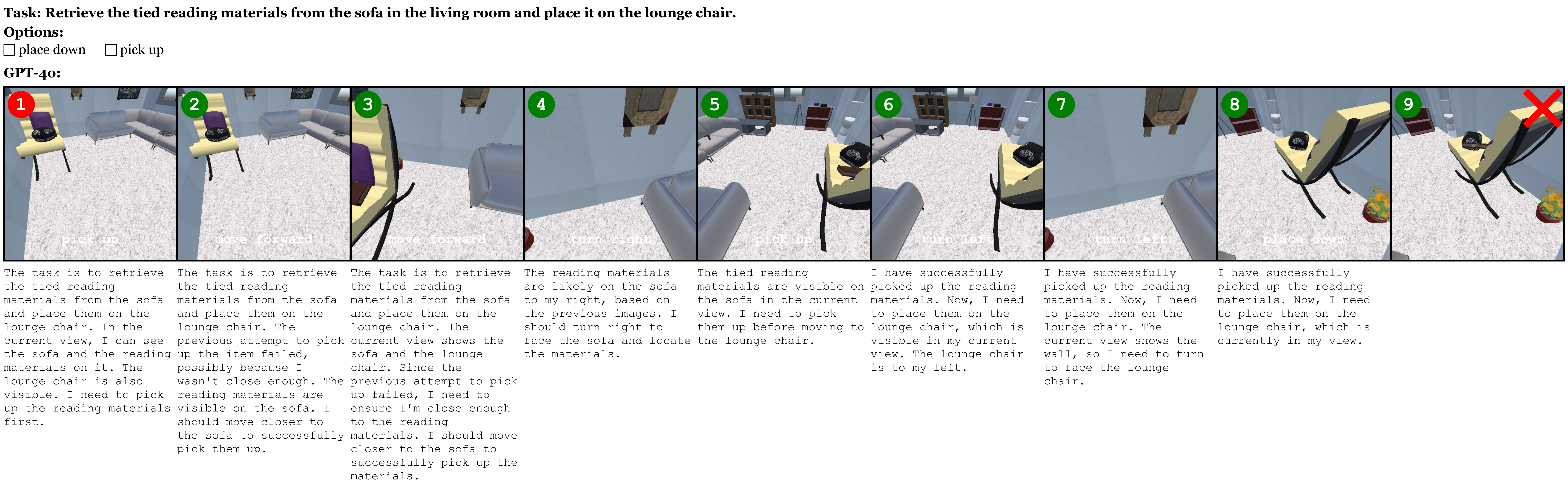}
  \vspace{-0.5em}
  \caption{}
  \label{fig:traj0028}
\end{figure*}

\subsection{Social Interaction}

\begin{figure*}[htbp]
  \centering
  \includegraphics[width=0.9\linewidth]{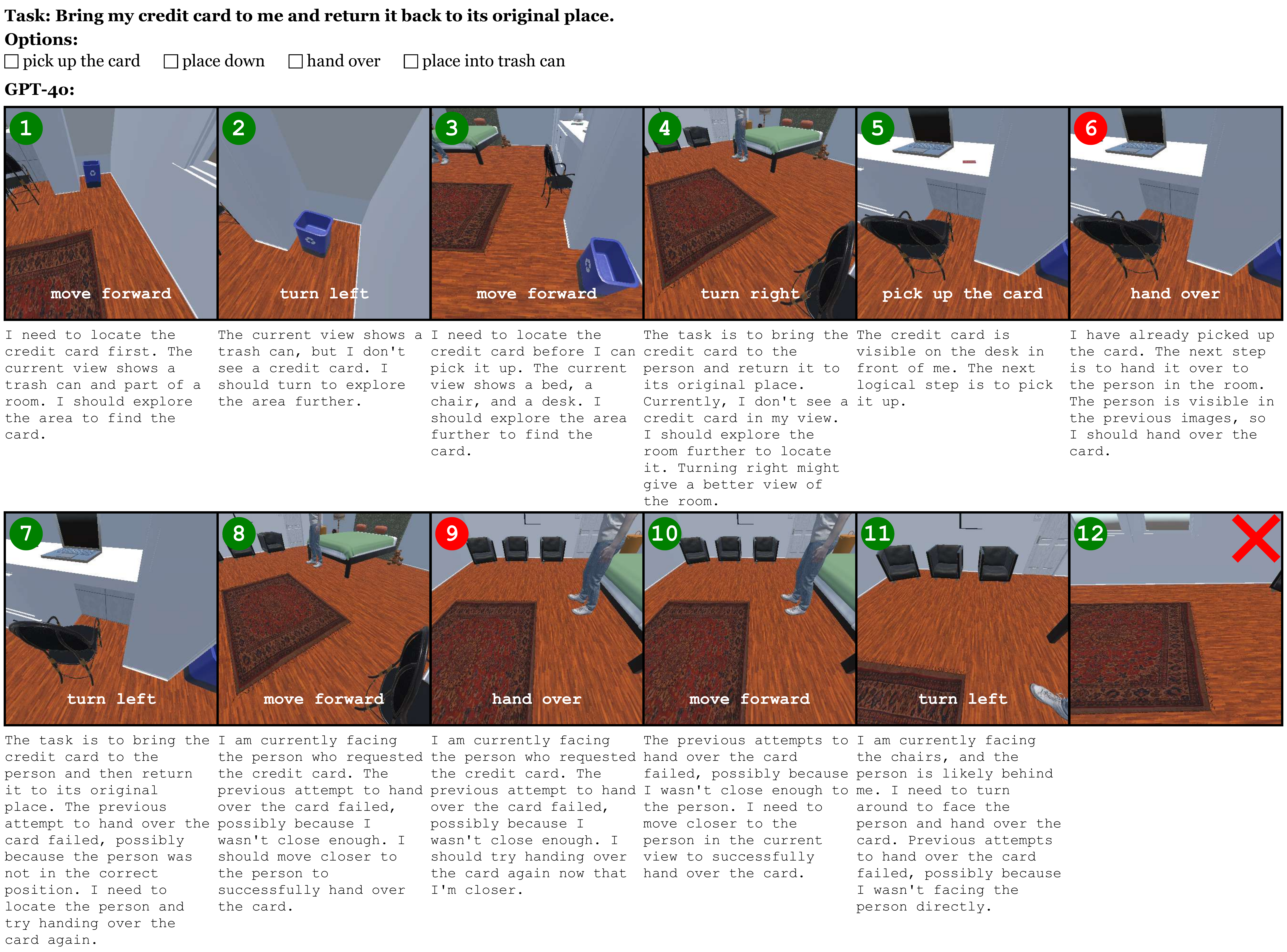}
  \vspace{-0.5em}
  \caption{}
  \label{fig:traj0018}
\end{figure*}

%% file: acl_latex.bbl
\begin{thebibliography}{79}
\providecommand{\natexlab}[1]{#1}

\bibitem[{Ahn et~al.(2022)Ahn, Brohan, Brown, Chebotar, Cortes, David, Finn, Fu, Gopalakrishnan, Hausman et~al.}]{ahn2022can}
Michael Ahn, Anthony Brohan, Noah Brown, Yevgen Chebotar, Omar Cortes, Byron David, Chelsea Finn, Chuyuan Fu, Keerthana Gopalakrishnan, Karol Hausman, et~al. 2022.
\newblock Do as i can, not as i say: Grounding language in robotic affordances.
\newblock \emph{arXiv preprint arXiv:2204.01691}.

\bibitem[{Anderson et~al.(2018{\natexlab{a}})Anderson, Chang, Chaplot, Dosovitskiy, Gupta, Koltun, Kosecka, Malik, Mottaghi, Savva et~al.}]{anderson2018evaluationembodiednavigationagents}
Peter Anderson, Angel Chang, Devendra~Singh Chaplot, Alexey Dosovitskiy, Saurabh Gupta, Vladlen Koltun, Jana Kosecka, Jitendra Malik, Roozbeh Mottaghi, Manolis Savva, et~al. 2018{\natexlab{a}}.
\newblock On evaluation of embodied navigation agents.
\newblock \emph{arXiv preprint arXiv:1807.06757}.

\bibitem[{Anderson et~al.(2018{\natexlab{b}})Anderson, Wu, Teney, Bruce, Johnson, S{\"u}nderhauf, Reid, Gould, and Van Den~Hengel}]{anderson2018vision}
Peter Anderson, Qi~Wu, Damien Teney, Jake Bruce, Mark Johnson, Niko S{\"u}nderhauf, Ian Reid, Stephen Gould, and Anton Van Den~Hengel. 2018{\natexlab{b}}.
\newblock Vision-and-language navigation: Interpreting visually-grounded navigation instructions in real environments.
\newblock In \emph{Proceedings of the IEEE conference on computer vision and pattern recognition}, pages 3674--3683.

\bibitem[{Bai et~al.(2023)Bai, Bai, Yang, Wang, Tan, Wang, Lin, Zhou, and Zhou}]{Qwen-VL}
Jinze Bai, Shuai Bai, Shusheng Yang, Shijie Wang, Sinan Tan, Peng Wang, Junyang Lin, Chang Zhou, and Jingren Zhou. 2023.
\newblock Qwen-vl: A versatile vision-language model for understanding, localization, text reading, and beyond.
\newblock \emph{arXiv preprint arXiv:2308.12966}.

\bibitem[{Batra et~al.(2020{\natexlab{a}})Batra, Chang, Chernova, Davison, Deng, Koltun, Levine, Malik, Mordatch, Mottaghi et~al.}]{batra2020rearrangement}
Dhruv Batra, Angel~X Chang, Sonia Chernova, Andrew~J Davison, Jia Deng, Vladlen Koltun, Sergey Levine, Jitendra Malik, Igor Mordatch, Roozbeh Mottaghi, et~al. 2020{\natexlab{a}}.
\newblock Rearrangement: A challenge for embodied ai.
\newblock \emph{arXiv preprint arXiv:2011.01975}.

\bibitem[{Batra et~al.(2020{\natexlab{b}})Batra, Gokaslan, Kembhavi, Maksymets, Mottaghi, Savva, Toshev, and Wijmans}]{batra2020objectnav}
Dhruv Batra, Aaron Gokaslan, Aniruddha Kembhavi, Oleksandr Maksymets, Roozbeh Mottaghi, Manolis Savva, Alexander Toshev, and Erik Wijmans. 2020{\natexlab{b}}.
\newblock Objectnav revisited: On evaluation of embodied agents navigating to objects.
\newblock \emph{arXiv preprint arXiv:2006.13171}.

\bibitem[{Chen et~al.(2024)Chen, Xu, Kirmani, Ichter, Sadigh, Guibas, and Xia}]{chen2024spatialvlm}
Boyuan Chen, Zhuo Xu, Sean Kirmani, Brain Ichter, Dorsa Sadigh, Leonidas Guibas, and Fei Xia. 2024.
\newblock Spatialvlm: Endowing vision-language models with spatial reasoning capabilities.
\newblock In \emph{Proceedings of the IEEE/CVF Conference on Computer Vision and Pattern Recognition}, pages 14455--14465.

\bibitem[{Chen et~al.(2023{\natexlab{a}})Chen, Li, Dong, Zhang, He, Wang, Zhao, and Lin}]{chen2023sharegpt4v}
Lin Chen, Jisong Li, Xiaoyi Dong, Pan Zhang, Conghui He, Jiaqi Wang, Feng Zhao, and Dahua Lin. 2023{\natexlab{a}}.
\newblock Sharegpt4v: Improving large multi-modal models with better captions.
\newblock \emph{arXiv preprint arXiv:2311.12793}.

\bibitem[{Chen et~al.(2023{\natexlab{b}})Chen, Ge, Ge, Ding, Li, Wang, Xu, Shan, and Liu}]{chen2023egoplan}
Yi~Chen, Yuying Ge, Yixiao Ge, Mingyu Ding, Bohao Li, Rui Wang, Ruifeng Xu, Ying Shan, and Xihui Liu. 2023{\natexlab{b}}.
\newblock Egoplan-bench: Benchmarking multimodal large language models for human-level planning.
\newblock \emph{arXiv preprint arXiv:2312.06722}.

\bibitem[{Cheng et~al.(2023)Cheng, Guo, Wu, Fang, Li, Liu, and Liu}]{egothink}
Sijie Cheng, Zhicheng Guo, Jingwen Wu, Kechen Fang, Peng Li, Huaping Liu, and Yang Liu. 2023.
\newblock Can vision-language models think from a first-person perspective?
\newblock \emph{arXiv preprint arXiv:2311.15596}.

\bibitem[{Cheng et~al.(2024{\natexlab{a}})Cheng, Guo, Wu, Fang, Li, Liu, and Liu}]{cheng2024egothink}
Sijie Cheng, Zhicheng Guo, Jingwen Wu, Kechen Fang, Peng Li, Huaping Liu, and Yang Liu. 2024{\natexlab{a}}.
\newblock Egothink: Evaluating first-person perspective thinking capability of vision-language models.
\newblock In \emph{Proceedings of the IEEE/CVF Conference on Computer Vision and Pattern Recognition}, pages 14291--14302.

\bibitem[{Cheng et~al.(2024{\natexlab{b}})Cheng, Leng, Zhang, Xin, Li, Chen, Zhu, Zhang, Luo, Zhao et~al.}]{cheng2024videollama}
Zesen Cheng, Sicong Leng, Hang Zhang, Yifei Xin, Xin Li, Guanzheng Chen, Yongxin Zhu, Wenqi Zhang, Ziyang Luo, Deli Zhao, et~al. 2024{\natexlab{b}}.
\newblock Videollama 2: Advancing spatial-temporal modeling and audio understanding in video-llms.
\newblock \emph{arXiv preprint arXiv:2406.07476}.

\bibitem[{Cheng et~al.(2024{\natexlab{c}})Cheng, Wang, Hu, Hu, Liu, Tu, Li, Shi, Liu, and Sun}]{cheng2024legent}
Zhili Cheng, Zhitong Wang, Jinyi Hu, Shengding Hu, An~Liu, Yuge Tu, Pengkai Li, Lei Shi, Zhiyuan Liu, and Maosong Sun. 2024{\natexlab{c}}.
\newblock Legent: Open platform for embodied agents.
\newblock \emph{arXiv preprint arXiv:2404.18243}.

\bibitem[{Das et~al.(2018)Das, Datta, Gkioxari, Lee, Parikh, and Batra}]{das2018embodied}
Abhishek Das, Samyak Datta, Georgia Gkioxari, Stefan Lee, Devi Parikh, and Dhruv Batra. 2018.
\newblock Embodied question answering.
\newblock In \emph{Proceedings of the IEEE conference on computer vision and pattern recognition}, pages 1--10.

\bibitem[{Deitke et~al.(2023)Deitke, Schwenk, Salvador, Weihs, Michel, VanderBilt, Schmidt, Ehsani, Kembhavi, and Farhadi}]{deitke2022objaverseuniverseannotated3d}
Matt Deitke, Dustin Schwenk, Jordi Salvador, Luca Weihs, Oscar Michel, Eli VanderBilt, Ludwig Schmidt, Kiana Ehsani, Aniruddha Kembhavi, and Ali Farhadi. 2023.
\newblock Objaverse: A universe of annotated 3d objects.
\newblock In \emph{Proceedings of the IEEE/CVF Conference on Computer Vision and Pattern Recognition}, pages 13142--13153.

\bibitem[{Deitke et~al.(2022)Deitke, VanderBilt, Herrasti, Weihs, Ehsani, Salvador, Han, Kolve, Kembhavi, and Mottaghi}]{deitke2022️}
Matt Deitke, Eli VanderBilt, Alvaro Herrasti, Luca Weihs, Kiana Ehsani, Jordi Salvador, Winson Han, Eric Kolve, Aniruddha Kembhavi, and Roozbeh Mottaghi. 2022.
\newblock Procthor: Large-scale embodied ai using procedural generation.
\newblock \emph{Advances in Neural Information Processing Systems}, 35:5982--5994.

\bibitem[{Dorbala et~al.(2024)Dorbala, Goyal, Piramuthu, Johnston, Ghanadhan, and Manocha}]{dorbala2024s}
Vishnu~Sashank Dorbala, Prasoon Goyal, Robinson Piramuthu, Michael Johnston, Reza Ghanadhan, and Dinesh Manocha. 2024.
\newblock S-eqa: Tackling situational queries in embodied question answering.
\newblock \emph{arXiv preprint arXiv:2405.04732}.

\bibitem[{Driess et~al.(2023)Driess, Xia, Sajjadi, Lynch, Chowdhery, Ichter, Wahid, Tompson, Vuong, Yu et~al.}]{driess2023palm}
Danny Driess, Fei Xia, Mehdi~SM Sajjadi, Corey Lynch, Aakanksha Chowdhery, Brian Ichter, Ayzaan Wahid, Jonathan Tompson, Quan Vuong, Tianhe Yu, et~al. 2023.
\newblock Palm-e: An embodied multimodal language model.
\newblock \emph{arXiv preprint arXiv:2303.03378}.

\bibitem[{Fan(2019)}]{fan2019egovqa}
Chenyou Fan. 2019.
\newblock Egovqa-an egocentric video question answering benchmark dataset.
\newblock In \emph{Proceedings of the IEEE/CVF International Conference on Computer Vision Workshops}, pages 0--0.

\bibitem[{Fu et~al.(2023)Fu, Chen, Shen, Qin, Zhang, Lin, Yang, Zheng, Li, Sun, Wu, and Ji}]{fu2023mme}
Chaoyou Fu, Peixian Chen, Yunhang Shen, Yulei Qin, Mengdan Zhang, Xu~Lin, Jinrui Yang, Xiawu Zheng, Ke~Li, Xing Sun, Yunsheng Wu, and Rongrong Ji. 2023.
\newblock Mme: A comprehensive evaluation benchmark for multimodal large language models.
\newblock \emph{arXiv preprint arXiv:2306.13394}.

\bibitem[{Fu et~al.(2024)Fu, Dai, Luo, Li, Ren, Zhang, Wang, Zhou, Shen, Zhang, Chen, Li, Lin, Zhao, Li, Xu, Zheng, Chen, Ji, and Sun}]{fu2024videomme}
Chaoyou Fu, Yuhan Dai, Yongdong Luo, Lei Li, Shuhuai Ren, Renrui Zhang, Zihan Wang, Chenyu Zhou, Yunhang Shen, Mengdan Zhang, Peixian Chen, Yanwei Li, Shaohui Lin, Sirui Zhao, Ke~Li, Tong Xu, Xiawu Zheng, Enhong Chen, Rongrong Ji, and Xing Sun. 2024.
\newblock Video-mme: The first-ever comprehensive evaluation of multi-modal llms in video analysis.
\newblock \emph{arXiv preprint arXiv:2405.21075}.

\bibitem[{Gibson(1977)}]{gibson1977theory}
JJ~Gibson. 1977.
\newblock The theory of affordances.
\newblock \emph{Perceiving, acting and knowing: Towards an ecological psychology/Erlbaum}.

\bibitem[{Gordon et~al.(2018)Gordon, Kembhavi, Rastegari, Redmon, Fox, and Farhadi}]{gordon2018iqa}
Daniel Gordon, Aniruddha Kembhavi, Mohammad Rastegari, Joseph Redmon, Dieter Fox, and Ali Farhadi. 2018.
\newblock Iqa: Visual question answering in interactive environments.
\newblock In \emph{Proceedings of the IEEE conference on computer vision and pattern recognition}, pages 4089--4098.

\bibitem[{He et~al.(2024)He, Luo, Bai, Hu, Thai, Shen, Hu, Han, Huang, Zhang et~al.}]{he2024olympiadbench}
Chaoqun He, Renjie Luo, Yuzhuo Bai, Shengding Hu, Zhen~Leng Thai, Junhao Shen, Jinyi Hu, Xu~Han, Yujie Huang, Yuxiang Zhang, et~al. 2024.
\newblock Olympiadbench: A challenging benchmark for promoting agi with olympiad-level bilingual multimodal scientific problems.
\newblock \emph{arXiv preprint arXiv:2402.14008}.

\bibitem[{Hu et~al.(2023)Hu, Yao, Wang, Wang, Pan, Chen, Yu, Wu, Zhao, Zhang et~al.}]{hu2024large}
Jinyi Hu, Yuan Yao, Chongyi Wang, Shan Wang, Yinxu Pan, Qianyu Chen, Tianyu Yu, Hanghao Wu, Yue Zhao, Haoye Zhang, et~al. 2023.
\newblock Large multilingual models pivot zero-shot multimodal learning across languages.
\newblock \emph{arXiv preprint arXiv:2308.12038}.

\bibitem[{Islam et~al.(2024)Islam, Gladstone, Islam, and Iqbal}]{islam2024eqamx}
Md~Mofijul Islam, Alexi Gladstone, Riashat Islam, and Tariq Iqbal. 2024.
\newblock {EQA}-{MX}: Embodied question answering using multimodal expression.
\newblock In \emph{The Twelfth International Conference on Learning Representations}.

\bibitem[{Jain et~al.(2019)Jain, Magalhaes, Ku, Vaswani, Ie, and Baldridge}]{jain2019stay}
Vihan Jain, Gabriel Magalhaes, Alexander Ku, Ashish Vaswani, Eugene Ie, and Jason Baldridge. 2019.
\newblock Stay on the path: Instruction fidelity in vision-and-language navigation.
\newblock \emph{arXiv preprint arXiv:1905.12255}.

\bibitem[{Jia et~al.(2024)Jia, Wang, Tong, Zhu, and Zheng}]{jia-etal-2024-langsuit}
Zixia Jia, Mengmeng Wang, Baichen Tong, Song-Chun Zhu, and Zilong Zheng. 2024.
\newblock \href {https://doi.org/10.18653/v1/2024.findings-acl.879} {{L}ang{S}uit{\textperiodcentered}{E}: Planning, controlling and interacting with large language models in embodied text environments}.
\newblock In \emph{Findings of the Association for Computational Linguistics: ACL 2024}, pages 14778--14814, Bangkok, Thailand. Association for Computational Linguistics.

\bibitem[{Kant et~al.(2022)Kant, Ramachandran, Yenamandra, Gilitschenski, Batra, Szot, and Agrawal}]{kant2022housekeep}
Yash Kant, Arun Ramachandran, Sriram Yenamandra, Igor Gilitschenski, Dhruv Batra, Andrew Szot, and Harsh Agrawal. 2022.
\newblock Housekeep: Tidying virtual households using commonsense reasoning.
\newblock In \emph{European Conference on Computer Vision}, pages 355--373. Springer.

\bibitem[{Khanna et~al.(2024{\natexlab{a}})Khanna, Mao, Jiang, Haresh, Shacklett, Batra, Clegg, Undersander, Chang, and Savva}]{khanna2023habitatsyntheticscenesdataset}
Mukul Khanna, Yongsen Mao, Hanxiao Jiang, Sanjay Haresh, Brennan Shacklett, Dhruv Batra, Alexander Clegg, Eric Undersander, Angel~X Chang, and Manolis Savva. 2024{\natexlab{a}}.
\newblock Habitat synthetic scenes dataset (hssd-200): An analysis of 3d scene scale and realism tradeoffs for objectgoal navigation.
\newblock In \emph{Proceedings of the IEEE/CVF Conference on Computer Vision and Pattern Recognition}, pages 16384--16393.

\bibitem[{Khanna et~al.(2024{\natexlab{b}})Khanna, Ramrakhya, Chhablani, Yenamandra, Gervet, Chang, Kira, Chaplot, Batra, and Mottaghi}]{khanna2024goat}
Mukul Khanna, Ram Ramrakhya, Gunjan Chhablani, Sriram Yenamandra, Theophile Gervet, Matthew Chang, Zsolt Kira, Devendra~Singh Chaplot, Dhruv Batra, and Roozbeh Mottaghi. 2024{\natexlab{b}}.
\newblock Goat-bench: A benchmark for multi-modal lifelong navigation.
\newblock In \emph{Proceedings of the IEEE/CVF Conference on Computer Vision and Pattern Recognition}, pages 16373--16383.

\bibitem[{Kim et~al.(2023)Kim, Kim, Kim, Min, and Choi}]{kim2024contextawareplanningenvironmentawarememory}
Byeonghwi Kim, Jinyeon Kim, Yuyeong Kim, Cheolhong Min, and Jonghyun Choi. 2023.
\newblock Context-aware planning and environment-aware memory for instruction following embodied agents.
\newblock In \emph{Proceedings of the IEEE/CVF International Conference on Computer Vision}, pages 10936--10946.

\bibitem[{Kolve and et~al.(2017)}]{ai2thor}
Eric Kolve and et~al. 2017.
\newblock Ai2-thor: An interactive 3d environment for visual ai.
\newblock \emph{arXiv preprint arXiv:1712.05474}.

\bibitem[{Ku et~al.(2020)Ku, Anderson, Patel, Ie, and Baldridge}]{ku2020room}
Alexander Ku, Peter Anderson, Roma Patel, Eugene Ie, and Jason Baldridge. 2020.
\newblock Room-across-room: Multilingual vision-and-language navigation with dense spatiotemporal grounding.
\newblock \emph{arXiv preprint arXiv:2010.07954}.

\bibitem[{Li et~al.(2023{\natexlab{a}})Li, Wang, Wang, Ge, Ge, and Shan}]{li2023seed}
Bohao Li, Rui Wang, Guangzhi Wang, Yuying Ge, Yixiao Ge, and Ying Shan. 2023{\natexlab{a}}.
\newblock Seed-bench: Benchmarking multimodal llms with generative comprehension.
\newblock \emph{arXiv preprint arXiv:2307.16125}.

\bibitem[{Li et~al.(2023{\natexlab{b}})Li, Zhang, Wong, Gokmen, Srivastava, Mart{\'\i}n-Mart{\'\i}n, Wang, Levine, Lingelbach, Sun et~al.}]{li2023behavior}
Chengshu Li, Ruohan Zhang, Josiah Wong, Cem Gokmen, Sanjana Srivastava, Roberto Mart{\'\i}n-Mart{\'\i}n, Chen Wang, Gabrael Levine, Michael Lingelbach, Jiankai Sun, et~al. 2023{\natexlab{b}}.
\newblock Behavior-1k: A benchmark for embodied ai with 1,000 everyday activities and realistic simulation.
\newblock In \emph{Conference on Robot Learning}, pages 80--93. PMLR.

\bibitem[{Li et~al.(2024)Li, Zhao, Wang, Wang, Zhou, Srivastava, Gokmen, Lee, Li, Zhang et~al.}]{embodiedagentinterface}
Manling Li, Shiyu Zhao, Qineng Wang, Kangrui Wang, Yu~Zhou, Sanjana Srivastava, Cem Gokmen, Tony Lee, Li~Erran Li, Ruohan Zhang, et~al. 2024.
\newblock Embodied agent interface: Benchmarking llms for embodied decision making.
\newblock \emph{arXiv preprint arXiv:2410.07166}.

\bibitem[{Lin et~al.(2023)Lin, Zhu, Ye, Ning, Jin, and Yuan}]{lin2023videollava}
Bin Lin, Bin Zhu, Yang Ye, Munan Ning, Peng Jin, and Li~Yuan. 2023.
\newblock Video-llava: Learning united visual representation by alignment before projection.
\newblock \emph{arXiv preprint arXiv:2311.10122}.

\bibitem[{Lin et~al.(2024)Lin, Yin, Ping, Molchanov, Shoeybi, and Han}]{lin2024vila}
Ji~Lin, Hongxu Yin, Wei Ping, Pavlo Molchanov, Mohammad Shoeybi, and Song Han. 2024.
\newblock Vila: On pre-training for visual language models.
\newblock In \emph{Proceedings of the IEEE/CVF Conference on Computer Vision and Pattern Recognition}, pages 26689--26699.

\bibitem[{Liu et~al.(2024{\natexlab{a}})Liu, Li, Wu, and Lee}]{liu2024visual}
Haotian Liu, Chunyuan Li, Qingyang Wu, and Yong~Jae Lee. 2024{\natexlab{a}}.
\newblock Visual instruction tuning.
\newblock \emph{Advances in neural information processing systems}, 36.

\bibitem[{Liu et~al.(2024{\natexlab{b}})Liu, Yu, Zhang, Xu, Lei, Lai, Gu, Ding, Men, Yang, Zhang, Deng, Zeng, Du, Zhang, Shen, Zhang, Su, Sun, Huang, Dong, and Tang}]{liu2024agentbench}
Xiao Liu, Hao Yu, Hanchen Zhang, Yifan Xu, Xuanyu Lei, Hanyu Lai, Yu~Gu, Hangliang Ding, Kaiwen Men, Kejuan Yang, Shudan Zhang, Xiang Deng, Aohan Zeng, Zhengxiao Du, Chenhui Zhang, Sheng Shen, Tianjun Zhang, Yu~Su, Huan Sun, Minlie Huang, Yuxiao Dong, and Jie Tang. 2024{\natexlab{b}}.
\newblock \href {https://openreview.net/forum?id=zAdUB0aCTQ} {Agentbench: Evaluating {LLM}s as agents}.
\newblock In \emph{The Twelfth International Conference on Learning Representations}.

\bibitem[{Liu et~al.(2023{\natexlab{a}})Liu, Yu, Zhang, Xu, Lei, Lai, Gu, Ding, Men, Yang et~al.}]{liu2023agentbench}
Xiao Liu, Hao Yu, Hanchen Zhang, Yifan Xu, Xuanyu Lei, Hanyu Lai, Yu~Gu, Hangliang Ding, Kaiwen Men, Kejuan Yang, et~al. 2023{\natexlab{a}}.
\newblock Agentbench: Evaluating llms as agents.
\newblock \emph{arXiv preprint arXiv:2308.03688}.

\bibitem[{Liu et~al.(2023{\natexlab{b}})Liu, Duan, Zhang, Li, Zhang, Zhao, Yuan, Wang, He, Liu et~al.}]{MMBench}
Yuan Liu, Haodong Duan, Yuanhan Zhang, Bo~Li, Songyang Zhang, Wangbo Zhao, Yike Yuan, Jiaqi Wang, Conghui He, Ziwei Liu, et~al. 2023{\natexlab{b}}.
\newblock Mmbench: Is your multi-modal model an all-around player?
\newblock \emph{arXiv preprint arXiv:2307.06281}.

\bibitem[{Liu et~al.(2023{\natexlab{c}})Liu, Li, Yang, Li, Yin, Liu, Jin, and Bai}]{ocrbench}
Yuliang Liu, Zhang Li, Biao Yang, Chunyuan Li, Xucheng Yin, Cheng-lin Liu, Lianwen Jin, and Xiang Bai. 2023{\natexlab{c}}.
\newblock On the hidden mystery of ocr in large multimodal models.
\newblock \emph{arXiv preprint arXiv:2305.07895}.

\bibitem[{Liu et~al.(2024{\natexlab{c}})Liu, Zhu, Shi, Zhang, Lou, Yang, Xi, Cao, Gu, Li et~al.}]{liu2024nvila}
Zhijian Liu, Ligeng Zhu, Baifeng Shi, Zhuoyang Zhang, Yuming Lou, Shang Yang, Haocheng Xi, Shiyi Cao, Yuxian Gu, Dacheng Li, et~al. 2024{\natexlab{c}}.
\newblock Nvila: Efficient frontier visual language models.
\newblock \emph{arXiv preprint arXiv:2412.04468}.

\bibitem[{Liu et~al.(2024{\natexlab{d}})Liu, Dong, Liu, Hu, Lu, and Rao}]{liu2024oryx}
Zuyan Liu, Yuhao Dong, Ziwei Liu, Winston Hu, Jiwen Lu, and Yongming Rao. 2024{\natexlab{d}}.
\newblock Oryx mllm: On-demand spatial-temporal understanding at arbitrary resolution.
\newblock \emph{arXiv preprint arXiv:2409.12961}.

\bibitem[{Lu et~al.(2024)Lu, Bansal, Xia, Liu, Li, Hajishirzi, Cheng, Chang, Galley, and Gao}]{lu2024mathvista}
Pan Lu, Hritik Bansal, Tony Xia, Jiacheng Liu, Chunyuan Li, Hannaneh Hajishirzi, Hao Cheng, Kai-Wei Chang, Michel Galley, and Jianfeng Gao. 2024.
\newblock \href {https://openreview.net/forum?id=KUNzEQMWU7} {Mathvista: Evaluating mathematical reasoning of foundation models in visual contexts}.
\newblock In \emph{The Twelfth International Conference on Learning Representations}.

\bibitem[{Ma et~al.(2024)Ma, Dai, Mu, Wu, Wang, Chi, Fei, Zhang, and Liu}]{ma2024doze}
Ji~Ma, Hongming Dai, Yao Mu, Pengying Wu, Hao Wang, Xiaowei Chi, Yang Fei, Shanghang Zhang, and Chang Liu. 2024.
\newblock Doze: A dataset for open-vocabulary zero-shot object navigation in dynamic environments.
\newblock \emph{arXiv preprint arXiv:2402.19007}.

\bibitem[{Majumdar et~al.(2024)Majumdar, Ajay, Zhang, Putta, Yenamandra, Henaff, Silwal, Mcvay, Maksymets, Arnaud et~al.}]{majumdar2024openeqa}
Arjun Majumdar, Anurag Ajay, Xiaohan Zhang, Pranav Putta, Sriram Yenamandra, Mikael Henaff, Sneha Silwal, Paul Mcvay, Oleksandr Maksymets, Sergio Arnaud, et~al. 2024.
\newblock Openeqa: Embodied question answering in the era of foundation models.
\newblock In \emph{Proceedings of the IEEE/CVF Conference on Computer Vision and Pattern Recognition}, pages 16488--16498.

\bibitem[{Misra et~al.(2018)Misra, Bennett, Blukis, Niklasson, Shatkhin, and Artzi}]{misra2018mapping}
Dipendra Misra, Andrew Bennett, Valts Blukis, Eyvind Niklasson, Max Shatkhin, and Yoav Artzi. 2018.
\newblock Mapping instructions to actions in 3d environments with visual goal prediction.
\newblock \emph{arXiv preprint arXiv:1809.00786}.

\bibitem[{Mu et~al.(2024)Mu, Zhang, Hu, Wang, Ding, Jin, Wang, Dai, Qiao, and Luo}]{mu2024embodiedgpt}
Yao Mu, Qinglong Zhang, Mengkang Hu, Wenhai Wang, Mingyu Ding, Jun Jin, Bin Wang, Jifeng Dai, Yu~Qiao, and Ping Luo. 2024.
\newblock Embodiedgpt: Vision-language pre-training via embodied chain of thought.
\newblock \emph{Advances in Neural Information Processing Systems}, 36.

\bibitem[{OpenAI(2023)}]{openai2023gpt4v}
OpenAI. 2023.
\newblock \href {https://cdn.openai.com/papers/GPTV_System_Card.pdf} {Gpt-4v(ision) system card}.

\bibitem[{OpenAI(2024)}]{openai2024gpt4o}
OpenAI. 2024.
\newblock \href {https://openai.com/index/hello-gpt-4o/} {Hello gpt4-o}.

\bibitem[{OpenGVLab(2024)}]{InternVL2}
OpenGVLab. 2024.
\newblock \href {https://internvl.github.io/blog/2024-07-02-InternVL-2.0/} {Internvl2: Better than the best—expanding performance boundaries of open-source multimodal models with the progressive scaling strategy}.

\bibitem[{Peng et~al.(2023)Peng, Wang, Dong, Hao, Huang, Ma, and Wei}]{peng2023kosmos}
Zhiliang Peng, Wenhui Wang, Li~Dong, Yaru Hao, Shaohan Huang, Shuming Ma, and Furu Wei. 2023.
\newblock Kosmos-2: Grounding multimodal large language models to the world.
\newblock \emph{arXiv preprint arXiv:2306.14824}.

\bibitem[{Qi et~al.(2020{\natexlab{a}})Qi, Wu, Anderson, Wang, Wang, Shen, and Hengel}]{qi2020reverie}
Yuankai Qi, Qi~Wu, Peter Anderson, Xin Wang, William~Yang Wang, Chunhua Shen, and Anton van~den Hengel. 2020{\natexlab{a}}.
\newblock Reverie: Remote embodied visual referring expression in real indoor environments.
\newblock In \emph{Proceedings of the IEEE/CVF Conference on Computer Vision and Pattern Recognition}, pages 9982--9991.

\bibitem[{Qi et~al.(2020{\natexlab{b}})Qi, Wu, Anderson, Wang, Wang, Shen, and van~den Hengel}]{qi2020reverieremoteembodiedvisual}
Yuankai Qi, Qi~Wu, Peter Anderson, Xin Wang, William~Yang Wang, Chunhua Shen, and Anton van~den Hengel. 2020{\natexlab{b}}.
\newblock \href {https://arxiv.org/abs/1904.10151} {Reverie: Remote embodied visual referring expression in real indoor environments}.
\newblock \emph{Preprint}, arXiv:1904.10151.

\bibitem[{Ren et~al.(2024)Ren, Clark, Dixit, Itkina, Majumdar, and Sadigh}]{ren2024explore}
Allen~Z Ren, Jaden Clark, Anushri Dixit, Masha Itkina, Anirudha Majumdar, and Dorsa Sadigh. 2024.
\newblock Explore until confident: Efficient exploration for embodied question answering.
\newblock \emph{arXiv preprint arXiv:2403.15941}.

\bibitem[{Savva and et~al.(2019)}]{habitat_challenge}
Manolis Savva and et~al. 2019.
\newblock Habitat challenge: A photorealistic embodied ai benchmark.
\newblock In \emph{Proceedings of the IEEE International Conference on Computer Vision (ICCV)}, pages 9338--9346. IEEE.

\bibitem[{Shridhar et~al.(2020)Shridhar, Thomason, Gordon, Bisk, Han, Mottaghi, Zettlemoyer, and Fox}]{shridhar2020alfred}
Mohit Shridhar, Jesse Thomason, Daniel Gordon, Yonatan Bisk, Winson Han, Roozbeh Mottaghi, Luke Zettlemoyer, and Dieter Fox. 2020.
\newblock Alfred: A benchmark for interpreting grounded instructions for everyday tasks.
\newblock In \emph{Proceedings of the IEEE/CVF conference on computer vision and pattern recognition}, pages 10740--10749.

\bibitem[{Singh et~al.(2019)Singh, Natarjan, Shah, Jiang, Chen, Parikh, and Rohrbach}]{textvqa}
Amanpreet Singh, Vivek Natarjan, Meet Shah, Yu~Jiang, Xinlei Chen, Devi Parikh, and Marcus Rohrbach. 2019.
\newblock Towards vqa models that can read.
\newblock In \emph{Proceedings of the IEEE Conference on Computer Vision and Pattern Recognition}, pages 8317--8326.

\bibitem[{Srivastava et~al.(2022)Srivastava, Li, Lingelbach, Mart{\'\i}n-Mart{\'\i}n, Xia, Vainio, Lian, Gokmen, Buch, Liu et~al.}]{srivastava2022behavior}
Sanjana Srivastava, Chengshu Li, Michael Lingelbach, Roberto Mart{\'\i}n-Mart{\'\i}n, Fei Xia, Kent~Elliott Vainio, Zheng Lian, Cem Gokmen, Shyamal Buch, Karen Liu, et~al. 2022.
\newblock Behavior: Benchmark for everyday household activities in virtual, interactive, and ecological environments.
\newblock In \emph{Conference on robot learning}, pages 477--490. PMLR.

\bibitem[{Szot et~al.(2023)Szot, Schwarzer, Agrawal, Mazoure, Metcalf, Talbott, Mackraz, Hjelm, and Toshev}]{szot2023large}
Andrew Szot, Max Schwarzer, Harsh Agrawal, Bogdan Mazoure, Rin Metcalf, Walter Talbott, Natalie Mackraz, R~Devon Hjelm, and Alexander~T Toshev. 2023.
\newblock Large language models as generalizable policies for embodied tasks.
\newblock In \emph{The Twelfth International Conference on Learning Representations}.

\bibitem[{Tan et~al.(2023)Tan, Ge, Guo, Liu, and Sun}]{tan2023knowledge}
Sinan Tan, Mengmeng Ge, Di~Guo, Huaping Liu, and Fuchun Sun. 2023.
\newblock Knowledge-based embodied question answering.
\newblock \emph{IEEE Transactions on Pattern Analysis and Machine Intelligence}, 45(10):11948--11960.

\bibitem[{Team et~al.(2023)Team, Anil, Borgeaud, Wu, Alayrac, Yu, Soricut, Schalkwyk, Dai, Hauth et~al.}]{team2023gemini}
Gemini Team, Rohan Anil, Sebastian Borgeaud, Yonghui Wu, Jean-Baptiste Alayrac, Jiahui Yu, Radu Soricut, Johan Schalkwyk, Andrew~M Dai, Anja Hauth, et~al. 2023.
\newblock Gemini: a family of highly capable multimodal models.
\newblock \emph{arXiv preprint arXiv:2312.11805}.

\bibitem[{Wei et~al.(2022)Wei, Wang, Schuurmans, Bosma, Xia, Chi, Le, Zhou et~al.}]{wei2022chain}
Jason Wei, Xuezhi Wang, Dale Schuurmans, Maarten Bosma, Fei Xia, Ed~Chi, Quoc~V Le, Denny Zhou, et~al. 2022.
\newblock Chain-of-thought prompting elicits reasoning in large language models.
\newblock \emph{Advances in neural information processing systems}, 35:24824--24837.

\bibitem[{Weihs et~al.(2021)Weihs, Deitke, Kembhavi, and Mottaghi}]{weihs2021visual}
Luca Weihs, Matt Deitke, Aniruddha Kembhavi, and Roozbeh Mottaghi. 2021.
\newblock Visual room rearrangement.
\newblock In \emph{Proceedings of the IEEE/CVF conference on computer vision and pattern recognition}, pages 5922--5931.

\bibitem[{Yang et~al.(2024{\natexlab{a}})Yang, Yang, Gupta, Han, Fei-Fei, and Xie}]{yang2024thinking}
Jihan Yang, Shusheng Yang, Anjali~W Gupta, Rilyn Han, Li~Fei-Fei, and Saining Xie. 2024{\natexlab{a}}.
\newblock Thinking in space: How multimodal large language models see, remember, and recall spaces.
\newblock \emph{arXiv preprint arXiv:2412.14171}.

\bibitem[{Yang et~al.(2024{\natexlab{b}})Yang, Sun, Weihs, VanderBilt, Herrasti, Han, Wu, Haber, Krishna, Liu, Callison-Burch, Yatskar, Kembhavi, and Clark}]{yang2024holodecklanguageguidedgeneration}
Yue Yang, Fan-Yun Sun, Luca Weihs, Eli VanderBilt, Alvaro Herrasti, Winson Han, Jiajun Wu, Nick Haber, Ranjay Krishna, Lingjie Liu, Chris Callison-Burch, Mark Yatskar, Aniruddha Kembhavi, and Christopher Clark. 2024{\natexlab{b}}.
\newblock \href {https://arxiv.org/abs/2312.09067} {Holodeck: Language guided generation of 3d embodied ai environments}.
\newblock \emph{Preprint}, arXiv:2312.09067.

\bibitem[{Yao et~al.(2022)Yao, Zhao, Yu, Du, Shafran, Narasimhan, and Cao}]{yao2022react}
Shunyu Yao, Jeffrey Zhao, Dian Yu, Nan Du, Izhak Shafran, Karthik Narasimhan, and Yuan Cao. 2022.
\newblock React: Synergizing reasoning and acting in language models.
\newblock \emph{arXiv preprint arXiv:2210.03629}.

\bibitem[{Yenamandra et~al.(2023)Yenamandra, Ramachandran, Yadav, Wang, Khanna, Gervet, Yang, Jain, Clegg, Turner et~al.}]{yenamandra2023homerobot}
Sriram Yenamandra, Arun Ramachandran, Karmesh Yadav, Austin Wang, Mukul Khanna, Theophile Gervet, Tsung-Yen Yang, Vidhi Jain, Alexander~William Clegg, John Turner, et~al. 2023.
\newblock Homerobot: Open-vocabulary mobile manipulation.
\newblock \emph{arXiv preprint arXiv:2306.11565}.

\bibitem[{You et~al.(2023)You, Zhang, Gan, Du, Zhang, Wang, Cao, Chang, and Yang}]{you2024ferret}
Haoxuan You, Haotian Zhang, Zhe Gan, Xianzhi Du, Bowen Zhang, Zirui Wang, Liangliang Cao, Shih-Fu Chang, and Yinfei Yang. 2023.
\newblock Ferret: Refer and ground anything anywhere at any granularity.
\newblock \emph{arXiv preprint arXiv:2310.07704}.

\bibitem[{Yu et~al.(2019)Yu, Chen, Gkioxari, Bansal, Berg, and Batra}]{yu2019multi}
Licheng Yu, Xinlei Chen, Georgia Gkioxari, Mohit Bansal, Tamara~L Berg, and Dhruv Batra. 2019.
\newblock Multi-target embodied question answering.
\newblock In \emph{Proceedings of the IEEE/CVF Conference on Computer Vision and Pattern Recognition}, pages 6309--6318.

\bibitem[{Yu et~al.(2024{\natexlab{a}})Yu, Yao, Zhang, He, Han, Cui, Hu, Liu, Zheng, Sun et~al.}]{yu2023rlhf}
Tianyu Yu, Yuan Yao, Haoye Zhang, Taiwen He, Yifeng Han, Ganqu Cui, Jinyi Hu, Zhiyuan Liu, Hai-Tao Zheng, Maosong Sun, et~al. 2024{\natexlab{a}}.
\newblock Rlhf-v: Towards trustworthy mllms via behavior alignment from fine-grained correctional human feedback.
\newblock In \emph{Proceedings of the IEEE/CVF Conference on Computer Vision and Pattern Recognition}, pages 13807--13816.

\bibitem[{Yu et~al.(2024{\natexlab{b}})Yu, Zhang, Yao, Dang, Chen, Lu, Cui, He, Liu, Chua, and Sun}]{yu2024rlaifv}
Tianyu Yu, Haoye Zhang, Yuan Yao, Yunkai Dang, Da~Chen, Xiaoman Lu, Ganqu Cui, Taiwen He, Zhiyuan Liu, Tat-Seng Chua, and Maosong Sun. 2024{\natexlab{b}}.
\newblock Rlaif-v: Aligning mllms through open-source ai feedback for super gpt-4v trustworthiness.
\newblock \emph{arXiv preprint arXiv:2405.17220}.

\bibitem[{Yue et~al.(2024)Yue, Ni, Zhang, Zheng, Liu, Zhang, Stevens, Jiang, Ren, Sun, Wei, Yu, Yuan, Sun, Yin, Zheng, Yang, Liu, Huang, Sun, Su, and Chen}]{yue2023mmmu}
Xiang Yue, Yuansheng Ni, Kai Zhang, Tianyu Zheng, Ruoqi Liu, Ge~Zhang, Samuel Stevens, Dongfu Jiang, Weiming Ren, Yuxuan Sun, Cong Wei, Botao Yu, Ruibin Yuan, Renliang Sun, Ming Yin, Boyuan Zheng, Zhenzhu Yang, Yibo Liu, Wenhao Huang, Huan Sun, Yu~Su, and Wenhu Chen. 2024.
\newblock Mmmu: A massive multi-discipline multimodal understanding and reasoning benchmark for expert agi.
\newblock In \emph{Proceedings of CVPR}.

\bibitem[{Zhang et~al.(2024{\natexlab{a}})Zhang, Li, Liu, Lee, Gui, Fu, Feng, Liu, and Li}]{zhang2024llava}
Yuanhan Zhang, Bo~Li, Haotian Liu, Yong~Jae Lee, Liangke Gui, Di~Fu, Jiashi Feng, Ziwei Liu, and Chunyuan Li. 2024{\natexlab{a}}.
\newblock \href {https://llava-vl.github.io/blog/2024-04-30-llava-next-video/} {Llava-next: A strong zero-shot video understanding model}.

\bibitem[{Zhang et~al.(2024{\natexlab{b}})Zhang, Zhu, Li, Liu, Ma, Chen, Jia, Huang, and Li}]{zhang2024taskorientedsequentialgrounding3d}
Zhuofan Zhang, Ziyu Zhu, Pengxiang Li, Tengyu Liu, Xiaojian Ma, Yixin Chen, Baoxiong Jia, Siyuan Huang, and Qing Li. 2024{\natexlab{b}}.
\newblock Task-oriented sequential grounding in 3d scenes.
\newblock \emph{arXiv preprint arXiv:2408.04034}.

\bibitem[{Zhu et~al.(2021)Zhu, Liang, Zhu, Yu, Chang, and Liang}]{zhu2021soon}
Fengda Zhu, Xiwen Liang, Yi~Zhu, Qizhi Yu, Xiaojun Chang, and Xiaodan Liang. 2021.
\newblock Soon: Scenario oriented object navigation with graph-based exploration.
\newblock In \emph{Proceedings of the IEEE/CVF Conference on Computer Vision and Pattern Recognition}, pages 12689--12699.

\end{thebibliography}
